\documentclass[11pt]{article}

\usepackage[preprint]{acl}

\usepackage{times}
\usepackage{latexsym}
\usepackage[T1]{fontenc}
\usepackage[utf8]{inputenc}
\usepackage{microtype}
\usepackage{inconsolata}
\usepackage{graphicx}
\usepackage{amsmath, amssymb, amsthm}
\usepackage{bm}   
\usepackage{booktabs}
\usepackage{multirow}
\usepackage{makecell}
\usepackage{array}
\usepackage{float} 
\usepackage{xcolor}
\usepackage[most]{tcolorbox}
\usepackage[table]{xcolor}
\usepackage[T1]{fontenc}
\usepackage[utf8]{inputenc}

\usepackage{microtype}
\usepackage{dsfont}
\usepackage{inconsolata}
\newcommand{\W}{\mathbf{W}} 
\usepackage{graphicx}
\newcommand{\methodname}{\textsc{CAGE-Cal}}
\newcommand{\select}{\textsc{CAGE-Select}}
\definecolor{promptframe1}{HTML}{927AB8}
\definecolor{promptframe2}{HTML}{5A87B8}
\newtcolorbox{promptbox}[1]{
  enhanced,
  colback=promptframe1!5,
  colframe=promptframe1,
  coltitle=white,
  fonttitle=\bfseries,
  title=#1,
  arc=1.5mm,
  boxrule=0.8pt,
  left=6pt, right=6pt, top=4pt, bottom=4pt,
}
\newtcolorbox{promptbox2}[1]{
  enhanced,
  colback=promptframe2!5,
  colframe=promptframe2,
  coltitle=white,
  fonttitle=\bfseries,
  title=#1,
  arc=1.5mm,
  boxrule=0.8pt,
  left=6pt, right=6pt, top=4pt, bottom=4pt,
}
\title{Counterfactual Graph for Multi-Agent LLM Calibration}

\author{
    Jiatan Huang\textsuperscript{1}, Mingchen Li\textsuperscript{2}, Ziming Li\textsuperscript{1}, Sunjae Kwon\textsuperscript{2}, Hong Yu\textsuperscript{2},Chuxu Zhang\textsuperscript{1,$\dag$} \\
    \textsuperscript{1}University of Connecticut~\textsuperscript{2}University of Massachusetts, Amherst\\
    \{jiatan.huang, chuxu.zhang\}@uconn.edu,
\textsuperscript{$\dag$}Corresponding author
}
\begin{document}
\maketitle
\begin{abstract}
Multi-agent LLM systems often treat agreement as evidence: when many agents in a panel give the same answer, that answer is assumed to be more reliable. 
We show that this assumption can fail after agents communicate. 
Communication can induce correlated failures and false consensus, so the same vote share may reflect reliable agreement in one topology but over-confidence in another.
We propose \methodname{}, 
a counterfactual agent-graph calibration framework for multi-agent LLMs. 
For each query, \methodname{} compares an observed post-communication agent graph with a matched counterfactual no-communication graph, capturing both pairwise failure correlations and group-level dependencies.
Rather than simply counting how many agents agree, \methodname{} estimates the counterfactual shift between observed and no-communication dependence, and calibrates confidence accordingly.
Across five benchmarks, \methodname{} improves reliability discrimination with competitive ECE, and its calibrated confidence further improves topology selection over the best fixed-topology strategy. Our code repo is available \href{https://anonymous.4open.science/r/Counterfactual-Graph-Calibration-for-Multi-Agent-LLMs-FDB0}{here}.
\end{abstract}

\section{Introduction}
\label{sec:intro}

% Multi-agent LLM systems are increasingly used in agentic pipelines such as clinical triage, code review, and search augmentation~\citep{du2023improvingfactualityreasoninglanguage,hong2024metagptmetaprogrammingmultiagent,wu2023autogenenablingnextgenllm}. These systems do not only produce answers. They also produce confidence estimates that decide how the answers are used. A high-confidence answer may be accepted automatically, while a low-confidence answer may be deferred to a human. Calibration is therefore not only an evaluation metric. It is a requirement for safe deployment~\cite{guo2017calibration,geifman2017selective}.
Multi-agent LLM systems are increasingly used in agentic pipelines such as code review, search augmentation, and clinical triage~\citep{du2023improvingfactualityreasoninglanguage,hong2024metagptmetaprogrammingmultiagent,wu2023autogenenablingnextgenllm}. 
In these settings, systems not only answer questions, but also indicate how much their answers should be trusted. This confidence signal directly affects downstream decisions: high-confidence answers may be accepted automatically, whereas low-confidence answers may be deferred to human review. Confidence should therefore be calibrated to the true probability of correctness, making calibration not merely an evaluation metric but a prerequisite for reliable and safe deployment~\citep{guo2017calibration,geifman2017selective}.

% Unlike single-agent calibration, which reflects the uncertainty of one model on one query, in the setting of multi-agent systems, raises a new class of
% challenges:
%Unlike single-agent calibration, which reflects the uncertainty of one model on one query~\cite{zhou2026steerconf,bani2026rewarding}, multi-agent calibration must account for a joint decision process among multiple agents whose outputs are shaped by communication topology. This setting raises a new class of challenges.
Unlike single-agent calibration, which characterizes the uncertainty of one model on one query~\citep{zhou2026steerconf,bani2026rewarding}, multi-agent calibration must account for a joint decision process in which multiple agents’ outputs are shaped by communication topology~\cite{huang2026evolverouter,zhang2025agentrouter,shi2026ng}. This setting introduces several distinct challenges.
1) \textit{Dependence-blind calibration}. Existing multi-agent calibration methods typically rely on uncertainty scores derived from the aggregate system output, such as vote share, entropy, or semantic disagreement~\citep{wang2023selfconsistencyimproveschainthought,kadavath2022languagemodelsmostlyknow,feng2025rethinkingllmuncertaintymultiagent,jiang2026discouq,chen2026responsecountsquantifyinguncertainty}. By treating agent responses as independent samples and compressing disagreement into a single scalar, these methods discard the structural information needed to distinguish benign disagreement from harmful agreement.
 2) \textit{Heterogeneous-agent calibration}. Existing multi-agent calibration methods primarily study agents instantiated from a shared LLM policy~\citep{zhu2026demystifying,qiao2026epistemic}. In practice, however, panels\footnote{A panel refers to the group of agents whose responses are aggregated to produce a final answer.} may include different LLMs with distinct knowledge boundaries, confidence scales, and failure modes. Calibrating confidence for such heterogeneous panels remains underexplored.
3) \textit{Topology generalization}. Existing calibration methods are largely designed and validated in all-to-all communication multi-agent systems~\citep{zhu2026demystifying,qiao2026epistemic}. Practical multi-agent systems, however, may use diverse communication topologies, including independent voting, chains, hub-spoke structures, and trees. Whether calibration methods developed for debate-style agents generalize to these more diverse and structurally complex settings remains unclear.

\begin{figure}[!t]
  \centering
  \includegraphics[width=\columnwidth]{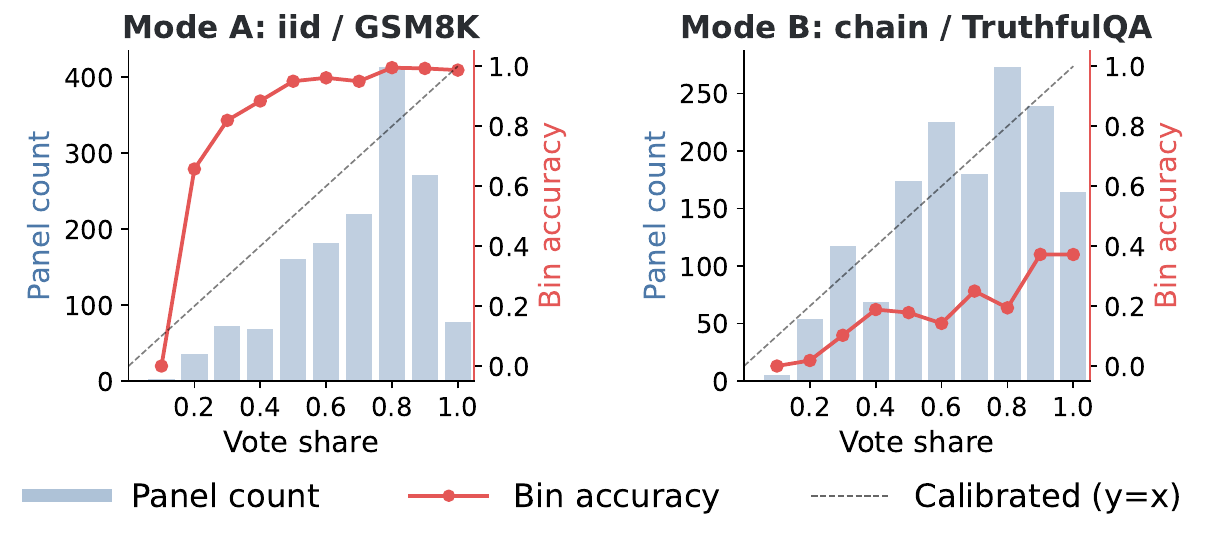}
  \vspace{-20pt}
  \caption{\textbf{The two failure modes.} 
Vote-share histogram (blue) vs.\ per-bin panel accuracy (red). 
The dashed line denotes perfect calibration. 
In iid/GSM8K, vote share underestimates accuracy (under confidence), while in chain/TruthfulQA, it overestimates accuracy (over confidence).}
  \label{fig:failure-modes}
  \vspace{-20pt}
\end{figure}
% To address these challenges, we first analyze multi-agent calibration across different LLMs and communication topologies, identifying two fundamental failure modes that cannot be captured by scalar disagreement scores.

To address these challenges, we first analyze how agreement and disagreement can become misleading confidence signals in both homogeneous and heterogeneous panels across different communication topologies. This analysis reveals two opposing failure modes.
(1) \emph{Diversity-induced under-confidence} (DUC): incorrect agents often spread across different wrong answers, so vote share underestimates panel reliability. (2) \emph{Communication-induced over-confidence} (COC): agents may converge to a wrong answer through influence, so vote share overestimates reliability. 

% These two cases can have similar answer distributions, but they require opposite calibration behavior.

Based on this diagnosis, we then propose \methodname{} 
(\textbf{C}ounterfactual \textbf{A}gent-\textbf{G}raph \textbf{Cal}ibration), 
a confidence calibration framework for heterogeneous multi-agent LLMs across diverse communication topologies.
For each query, \methodname{} compares the observed communication graph with an IID counterfactual graph, where the same agents vote independently without communication. Their difference captures communication-induced dependence, enabling the calibrator to down-calibrate COC cases where high agreement reflects agent coupling rather than independent evidence. To address DUC, \methodname{} further combines pairwise failure correlations with high-order hyperedges over model family, prompt role, answer cluster, and topology. These group-level signals reveal when disagreement is benign, e.g., when wrong agents are dispersed across weak answer groups.
Together, these components adjust confidence in opposite directions: lowering confidence for false consensus and preserving or increasing confidence for harmless diversity.

Our contributions are summarized as follows:
\begin{itemize}
    \item We analyze multi-agent calibration across different LLMs and communication topologies, identifying two failure modes: DUC and COC.
    \item We propose \methodname{}, a new multi-agent confidence calibration framework that calibrates confidence by contrasting communication graphs with IID counterfactual graphs and modeling failure dependencies.

    \item We evaluate \methodname{} on five benchmarks and  topologies, showing improved reliability discrimination with competitive calibration error. We further introduce \select{}, which uses calibrated confidence to select the most reliable topology  and improve final panel accuracy.
\end{itemize}
% \begin{itemize}
%     \item We introduce instance-conditional agent correlation graphs for studying multi-agent calibration. This representation captures how agent failures depend on the query, model family, and communication topology. Using it, we identify two failure modes that scalar disagreement scores miss: diversity-induced under-confidence and communication-induced over-confidence.

%     \item We propose CAGE-CAL, a counterfactual graph and hypergraph calibration framework for multi-agent LLMs. For each query, CAGE-CAL compares the observed post-communication graph with an iid counterfactual graph over the same matched agents. It models both pairwise failure correlations and higher-order dependency groups, including model family, prompting role, answer cluster, and communication topology.

%     \item We evaluate CAGE-CAL across five benchmarks and five communication topologies. CAGE-CAL improves reliability discrimination while maintaining competitive calibration error. We further show that its calibrated confidence can serve as a control signal through CAGE-Select, which improves final panel accuracy by selecting the most reliable topology per query.
% \end{itemize}
\section{Related Work}
\label{sec:related}
% Our work on multi-agent confidence calibration is situated at the intersection of two rapidly developing research areas: uncertainty quantification (UQ) for LLM-based agents and confidence calibration for LLMs.

% \paragraph{Multi-agent LLM systems and correlated errors.}
% \textbf{(why the need the calibation for mulati agent system)}
% Composing LLMs into multi-agent systems has improved factuality, reasoning, and code synthesis~\citep{du2023improvingfactualityreasoninglanguage,hong2024metagptmetaprogrammingmultiagent,wu2023autogenenablingnextgenllm}. 
% These gains are often attributed to diversity among agents, but this diversity is limited when agents share training data, model families, alignment procedures, or prompting patterns. 
% Recent work shows that LLM errors can be correlated across models~\citep{estornell2026multillmdebate}, especially within the same model family~\citep{kim2025correlatederrorslargelanguage}, and that consensus is not always reliable evidence when agents share failure modes~\citep{denisovblanch2026consensusverificationcrowdwisdom}. 
% We build on this observation and ask how multi-agent confidence should be calibrated when failures are correlated by design. 
% Our answer is to model the dependency structure among agents directly, rather than treating agreement as independent evidence.

\paragraph{Multi-Agent Uncertainty Quantification.}
Existing uncertainty estimators for LLM panels often summarize the answer distribution with a scalar score~\citep{wang2023selfconsistencyimproveschainthought,kadavath2022languagemodelsmostlyknow,feng2025rethinkingllmuncertaintymultiagent,chen2026responsecountsquantifyinguncertainty,jiang2026discouq}.  
For example, DiscoUQ-LLM~\cite{jiang2026discouq} learns from disagreement features, while MATU~\cite{chen2026responsecountsquantifyinguncertainty} models uncertainty through tensor decomposition over multi-agent rollouts~\citep{jiang2026discouq, chen2026responsecountsquantifyinguncertainty}. These methods are useful when disagreement reflects uncertainty. 
However, disagreement can be misleading: benign diversity may cause under-confidence, while communication-induced agreement may cause over-confidence. \methodname{}  therefore conditions confidence on agent dependencies rather than disagreement alone.
% Vote share underlies majority voting and self-consistency~\citep{wang2023selfconsistencyimproveschainthought}. 
% Mean log-probability extends single-model confidence estimation to the panel setting~\citep{kadavath2022languagemodelsmostlyknow}. 
% Entropy-based methods, such as DiverseAgentEntropy~\citep{feng2025rethinkingllmuncertaintymultiagent}, use disagreement among agents as a black-box uncertainty signal. 

% However, our results show that this assumption can fail in two directions: benign diversity can make a reliable panel look uncertain, while communication can make an unreliable panel look confident. 
% CAGE-CAL therefore conditions confidence on the dependency structure that produced the disagreement, rather than on the disagreement level alone.

\paragraph{Multi-Agent Confidence Calibration.}

% such as Platt scaling~\cite{platt1999probabilistic}, isotonic regression~\cite{zadrozny2002transforming}  scaling-binning~\cite{kumar2019verified} 

% Confidence calibration aims to align a model's predicted probability with the true likelihood of correctness.

Existing work can be broadly grouped into three lines. The first line, \textit{Post-hoc plurality calibrators}~\cite{zhang2026agentic}, calibrates scalar uncertainty signals, such as vote share or entropy, using post-hoc methods~\cite{platt1999probabilistic,zadrozny2002transforming,kumar2019verified}.
% The first line, \textit{post-hoc scalar calibration}~\cite{zhang2026agentic}, applies post-hoc methods\cite{platt1999probabilistic,zadrozny2002transforming,kumar2019verified} such as scaling-binning~\cite{kumar2019verified} to scalar uncertainty scores such as vote share and entropy,
% but has been mainly studied in single-agent settings.
%%
The second line studies \textit{LLM-elicited confidence estimators}~\cite{lin2022teaching,yang2024confidence}, where LLMs are trained or prompted to report calibrated confidence. For example, LLM-Cal~\cite{lin2022teaching} fine-tunes an LLM to align verbalized confidence with empirical correctness. 
The third line develops \textit{trained calibrators}~\cite{li2025graph,jiang2026discouq,zhu2026demystifying}, which estimate final-answer correctness from confidence and disagreement features. For example, DiscoUQ-LLM~\cite{jiang2026discouq} uses logistic regression over vote confidence and hand-crafted disagreement features. 
Despite their effectiveness, these methods largely overlook communication-induced dependencies, heterogeneous LLM panels, and diverse communication topologies.

%SCollaborative
% Calibration (Collab Cal)~\cite{yang2024confidence} further performs post-hoc, training-free calibration by using multiple tool-augmented LLM agents to generate answers, verbalized confidence, and feedback through a simulated group deliberation process.
% However, these methods primarily rely on verbalized uncertainty or deliberative feedback, and do not explicitly model how communication topology induces dependencies among agents.
%%
\section{Problem Statement}
\label{sec:setup}
% \paragraph{Task Statement.}
A multi-agent panel consists of $N$ language model agents, a communication topology represented by adjacency matrix $W_c$, and a communication protocol.
For a query $x$, each agent emits a final answer $y_i(x)$, and the panel produces one prediction
$
\hat{y}(x)=\mathrm{plurality}(\{y_i(x)\}_{i=1}^{N}),
$
where plurality vote selects the most frequent normalized answer.
Given the panel prediction $\hat{y}(x)$, our task is to estimate the calibrated probability that this panel answer is correct:
\begin{equation}
\hat{p}(x)
=
\Pr\big(C(x)=1 \mid \mathcal{M}(x), W_c\big),
\end{equation}
where
$
C(x)=\mathbf{1}[\hat{y}(x)=y^\star(x)]
$
denotes whether the panel prediction is correct. 
Here, $\mathcal{M}(x)$ denotes the observable final panel state, including agents' final answers, answer-level confidence scores, vote statistics, and answer embeddings.
For later use, we also define the agent-level correctness indicator
$
c_i(x)=\mathbf{1}[y_i(x)=y^\star(x)],
$
which denotes whether agent $i$ answers correctly.
% This setting separates reliability estimation from answer generation. 
% Two panels may produce the same answer with the same vote share, but their reliability can differ if the agreement is produced by different dependency structures.

% \paragraph{Instance-Conditional Agent Correlation Graphs}
% % Scalar uncertainty scores, such as vote share, answer entropy, and mean log probability, compress the panel into one number. 
% % This compression discards the structure needed to tell whether agreement is independent evidence or correlated failure. 

% To distinguish independent agreement from correlated failure,  we construct an instance-conditional agent correlation graph. In this graph, the vertices represent agents, while the weighted edges capture conditional dependencies between agents' correctness on queries similar to $x$: 
% $
% W_{ij}(x)=\mathrm{Corr}(c_i,c_j \mid q \approx x)
% $
% ,which measures whether agents $i$ and $j$ tend to succeed or fail together on queries similar to $x$. 
% % This local conditioning matters because agent dependence can change across factual, truthfulness, and reasoning queries.
% %
% We estimate $W(x)$ from the training split only. 
% For each validation or test query, we retrieve its $k$ nearest training queries in embedding space and compute the empirical Pearson correlation of each agent pair over this local neighborhood. 
% Validation and test labels are never used to estimate $W(x)$; they are used only to train, validate, or evaluate the calibrator.

% \input{sections/04_corrected_graph}
% \input{sections/05_correctness_shift}
\section{Calibration Failure Modes}
\label{sec:failure_modes}
To diagnose calibration failures in multi-agent systems, we analyze panel behavior across a grid of communication topologies and benchmarks 
% The study covers five topologies (iid, debate, chain, hub-spoke, and tree) and five benchmarks spanning knowledge, truthfulness, reasoning, and math. For each instance, we sample heterogeneous panels from a shared agent pool with variation in both backbone model family and prompting role, and evaluate each topology--benchmark cell over three rollouts. 
% This controlled design allows us to compare how the same pool of agents behaves under different communication structures.
and identify two recurring failure modes: Diversity-Induced Under-Confidence (DUC) and Communication-Induced Over-Confidence (COC). 
Figure~\ref{fig:failure-modes} illustrates these modes with two representative cells: iid on GSM8K and chain on TruthfulQA. 
We defer the full analysis pipeline to Appendix~\ref{app:failure_analysis}. 

\paragraph{Failure mode A: DUC.}

Let \(v_a(x)\) denote the fraction of agents that output answer \(a\) for query \(x\):
$
v_a(x)=\frac{1}{N}\sum_{i=1}^{N}\mathbf{1}[y_i(x)=a].
$
Let \(v(x)=v_{\hat{y}(x)}(x)\) be the vote share of the plurality answer \(\hat{y}(x)\). 
For a vote-share bin centered at \(s\), define
$
\mathcal{B}_s=\{x: v(x)\in[s-\epsilon,s+\epsilon]\}.
$
DUC occurs when examples in this bin are empirically more accurate than their vote share suggests, while competing answers remain weakly supported:
\begin{equation}
    \begin{aligned}
\frac{1}{|\mathcal{B}_s|}\sum_{x\in\mathcal{B}_s} C(x) &> s, \\
\frac{1}{|\mathcal{B}_s|}\sum_{x\in\mathcal{B}_s}
\max_{a\neq \hat{y}(x)} v_a(x) &\ll s.
\end{aligned}
\end{equation}

% The first condition captures under-confidence: among examples with plurality vote share near \(s\), the empirical correctness rate exceeds \(s\). 
% The second condition captures benign diversity: the largest competing answer receives much less support than the plurality answer, so wrong votes are dispersed rather than forming a strong alternative consensus.

% \paragraph{Pattern.} In high-accuracy cells, wrong agents often spread across different wrong answers. 
% Each wrong answer receives only a few votes, while the correct answer remains the plurality answer. 
% The panel is reliable, but its vote share can look modest. 
% A disagreement-based calibrator then mistakes benign diversity for uncertainty and assigns too little confidence.

% \paragraph{Cause.}
% This failure is not mainly caused by harmful communication. 
% It comes from the heterogeneous structure of the panel. 
% Different backbones and prompting roles can produce different wrong answers, so the final answer distribution looks dispersed even when the plurality answer is usually correct. 
% Thus, high answer entropy does not always mean high risk. 
% It can also mean that the wrong votes are weakly clustered and do not form a serious competing consensus.

\paragraph{Failure mode B: COC.}
COC occurs when examples in a high-vote-share bin are empirically less accurate than their vote share suggests, while the agents supporting the plurality answer are strongly dependent:
\begin{equation}
   \begin{aligned}
\frac{1}{|\mathcal{B}_s|}\sum_{x\in\mathcal{B}_s} C(x) &< s, \\
\frac{1}{|\mathcal{B}_s|}\sum_{x\in\mathcal{B}_s}
\mathrm{Dep}\big(S_{\hat{y}}(x)\big) &\gg 0,
\end{aligned} 
\end{equation}
where $S_{\hat{y}}(x)=\{i:y_i(x)=\hat{y}(x)\}$ is the set of agents supporting the plurality answer, and \(\mathrm{Dep}(S_{\hat{y}}(x))\) measures the average dependence among these supporting agents.

\begin{figure*}[t]
    \centering
    \includegraphics[width=\textwidth]{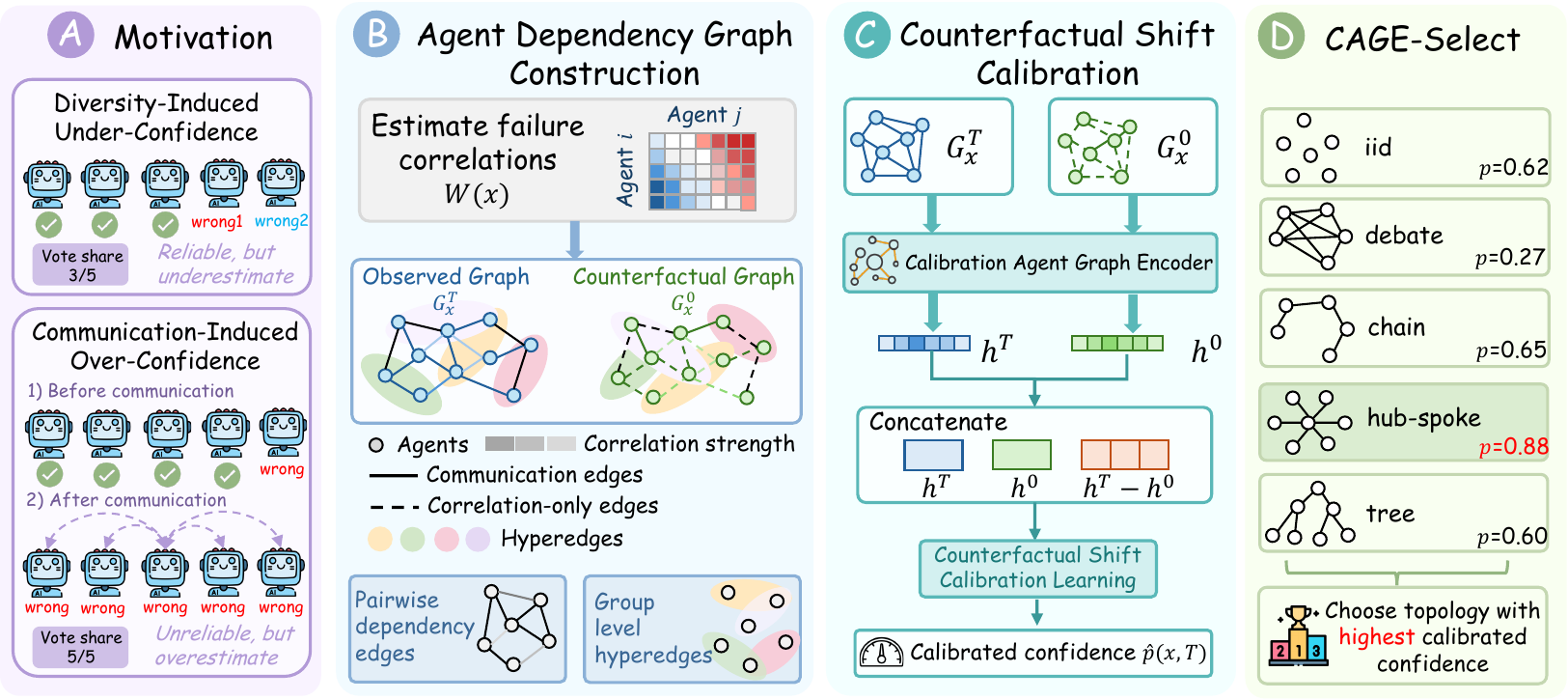}
    \vspace{-15pt}
    \caption{
    \textbf{Overview of \methodname{}.} 
    The same vote share can reflect different reliability depending on how agreement is formed. 
    \methodname{} constructs matched observed and IID counterfactual dependency graphs, encodes pairwise and group-level dependencies, and uses the counterfactual shift to estimate calibrated panel confidence. 
    The resulting confidence can also guide topology selection.
    }
    \vspace{-15pt}
    \label{fig:main}
\end{figure*}

\section{\methodname{}: A Counterfactual Graph Calibration Framework} %   Counterfactual Agent Graph Calibration}
\label{sec:method}
\subsection{Motivation and Approach Overview}
Based on the analysis in Section~\ref{sec:failure_modes}, we find that the same vote pattern can imply opposite reliability. 
Benign diversity disperses wrong votes and makes reliable panels appear uncertain, while communication can concentrate agents on the same wrong answer and create false confidence. 
This motivates \methodname{}, which calibrates confidence by modeling the dependency structure behind agent agreement. 
Figure~\ref{fig:main} (A) illustrates the two motivating failure modes.
% also found effective multi-agent confidence calibration requires going beyond the final answer distribution.
% % The same level of agreement can correspond to different reliability: diversity-induced disagreement can make a reliable panel appear uncertain, whereas communication-induced agreement can make an unreliable panel appear confident. 
% % Thus, the calibrator must model the dependency structure that gives rise to agreement or disagreement, rather than relying only on vote share or entropy.
% % \paragraph{Motivation.}

% To address this limitation, we propose \methodname{}. 

Figure~\ref{fig:main} (B-C) provides an overview of the \methodname{} framework, which consists of three components. 
First, \textit{Agent Dependency Graph Construction} builds two instance-level graphs: a counterfactual IID graph \(G_x^0\), which captures dependencies that would exist without communication, and a post-communication graph \(G_x^T\), which captures dependencies after agents interact under topology \(T\). 
Second, \textit{Counterfactual Shift Calibration}, which includes  1) \textit{Calibration Agent Graph Encoder} and 2) \textit{Counterfactual Shift Calibration Learning}, uses the difference between these two graphs as a calibration signal to determine whether agreement reflects independent evidence or correlated failure. 
Finally, we introduce \textsc{CAGE-Select} (D), which uses the calibrated confidence from \methodname{} to select the most reliable topology for each query and improve final panel accuracy.

\subsection{Agent Dependency Graph Construction}

For each query $x$ and communication topology $T$, \methodname{} constructs two graphs over matched agent identities. 
The observed graph $G_x^T$ is built from the completed panel under topology $T$. 
The counterfactual graph $G_x^0$ is built from the cached iid run of the same query and rollout, where the same agents answer without communication. 
We keep the shared agent identities between the two runs, so that the two graphs describe the same panel under two relational states. 
This matching makes the contrast between $G_x^T$ and $G_x^0$ reflect communication induced changes rather than changes in the agent population.

% Because the nodes are matched, differences between the two graphs reflect communication induced changes rather than changes in the agent population.
\paragraph{Node.}
Each node represents one agent. 
For both $G_x^T$ and $G_x^0$, we use the same 23-dimensional node feature schema:
\begin{equation}
v_i(x)
=
[
s_i,\,
p_i,\,
r_i,\,
\mu_i,\,
\sigma_i^2,\,
m_i,\,
n^{-1},\,
a_i
],
\end{equation}
where $s_i$ is the agent confidence score.  $p_i$ indicates whether the agent supports the panel plurality answer, $r_i$ is the normalized vote rank of the agent's answer, $(\mu_i,\sigma_i^2,m_i)$ are local correlation summaries from the corresponding row of the instance conditional correlation matrix, $n^{-1}$ is the inverse panel size, and $a_i$ is a low dimensional answer embedding.  Further details on node feature construction are provided in Appendix~\ref{con:details_of_node_feature}.

% The confidence score $s_i$ is computed from the agent's mean answer log probability. 
% We clip the mean log probability to $[-10,0]$ and linearly map it to a normalized score. 
% The plurality indicator $p_i$ is one when the agent's normalized answer matches the panel plurality answer. 
% The vote rank $r_i$ is normalized by the number of distinct answer clusters in the panel. 
% The correlation summaries $(\mu_i,\sigma_i^2,m_i)$ are the row mean, row variance, and row maximum of $|W_{ij}(x)|$ over other agents. 
% For the observed tower, these summaries are computed from the post communication correlation matrix. 
% For the iid counterfactual tower, they are computed from the iid correlation matrix. 
% Finally, the answer embedding $a_i$ is obtained by encoding the agent's answer with Sentence-BERT and reducing the embedding to 16 dimensions by PCA. 
% Thus, each node records both the agent's individual contribution to the vote and its local dependence pattern with the rest of the panel.

\paragraph{Pairwise dependency edges.}
Pairwise edges describe local dependence between agents. 
For agents $i$ and $j$, the edge feature in the observed graph is
$
e_{ij}^T(x)
=
[
W_{c,T}^{ij},\,
W_T^{ij}(x)
]
$
, where $W^{ij}_{c,T}$ is the binary communication adjacency under topology $T$ and records who could directly observe or influence whom under the topology. $W_T^{ij}(x)$ is the instance-conditional correctness correlation estimated from nearby training queries, indicating whether agents $i$ and $j$ tend to succeed or fail together on queries similar to $x$.  Further details about this correlation estimation are provided in Appendix~\ref{con:corrlation_extimtion}. 
In the IID counterfactual graph, the communication-adjacency feature is set to zero, since no communication occurs among agents.
% The iid counterfactual graph uses
% \[
% e_{ij}^0(x)
% =
% [
% 0,\,
% W_0^{ij}(x)
% ],
% \]
% because no agent communicates in the counterfactual iid run.

% The two edge dimensions play different roles. 
% The communication adjacency records who could directly observe or influence whom under the topology. 
% The local correlation term records which agents tend to succeed or fail together on queries similar to $x$. 

\paragraph{Group level hyperedges.}
Pairwise edges capture only dyadic dependence, while many sources of dependence act at the group level. 
We therefore augment each graph with a hyperedge set
\begin{equation}
   \mathcal{H}_x^T
=
\mathcal{H}_{\mathrm{fam}}
\cup
\mathcal{H}_{\mathrm{role}}
\cup
\mathcal{H}_{\mathrm{ans}}(x)
\cup
\mathcal{H}_{\mathrm{topo}}^T.
\end{equation}
Here, $\mathcal{H}_{\mathrm{fam}}$ connects agents sharing the same backbone family, 
$\mathcal{H}_{\mathrm{role}}$ connects agents using the same prompting role, 
$\mathcal{H}_{\mathrm{ans}}(x)$ connects agents that emit the same normalized answer for query $x$, 
and $\mathcal{H}_{\mathrm{topo}}^T$ connects agents exposed through topology $T$.
The IID counterfactual graph does not include this communication exposure hyperedge.
% Pairwise edges alone do not capture all sources of dependence. 
% Many dependencies act on groups of agents. 
% For this reason, each graph also includes hyperedges over four types of groups.
% First, model family hyperedges connect agents that share the same backbone family. 
% In our agent pool, these families include Qwen, Llama, Gemma, and Phi variants. 
% Second, prompting role hyperedges connect agents that use the same prompting role. 
% Third, answer cluster hyperedges connect agents that emit the same normalized answer within the panel. 
% The answer normalization lowercases the answer and removes surrounding punctuation and whitespace before exact string matching. 
% Fourth, for non-iid observed graphs, we add a topology exposure hyperedge over the agents that participate in the communication topology. 
We create a hyperedge only when it contains at least two agents. 
These hyperedges are dependence units rather than auxiliary metadata. 
A shared backbone can induce shared blind spots. 
A shared prompting role can induce similar reasoning patterns. 
An answer cluster can become a consensus unit. 
A communication exposure group can reflect shared influence from the topology. 
This design follows the empirical analysis in Appendix~\ref{con:analysis_progress}: multi-agent reliability is shaped by both population side structure and topology induced coupling~\cite{li2026graph}.
% \subsection{Counterfactual Shift Calibration}
\subsection{Calibration Agent Graph Encoder}
% \subsection{Encoding the Counterfactual Shift}
Given the matched graphs \(G_x^T\) and \(G_x^0\), \methodname{}
uses the calibration agent graph encoder \(f_\theta\) to obtain:
\begin{equation}
    h^T = \mathrm{Pool}\!\left(f_\theta(G_x^T)\right),
    \qquad
    h^0 = \mathrm{Pool}\!\left(f_\theta(G_x^0)\right).
\end{equation}
For each graph, the encoder updates every agent representation with two
relational signals. The pairwise update aggregates over agent-agent edges:
\begin{equation}
    p_i^{(\ell)}
    =
    \sigma\!\left(
    \sum_{j\in\mathcal{N}(i)}
    \alpha_{ij}^{(\ell)}
    W_p^{(\ell)} z_j^{(\ell)}
    \right),
\end{equation}
where \(\alpha_{ij}^{(\ell)}\) is the attention weight from agent \(j\) to
agent \(i\) at layer \(\ell\). It determines how much information agent
\(i\) receives from agent \(j\). The weight is computed using the two agent representations and their edge feature
\(e_{ij}(x)=[W_c^{ij}, W_{ij}(x)]\), allowing communication exposure and query-specific failure correlation to condition message passing. Here, \(z_i^{(\ell)}\) denotes the layer-\(\ell\)
representation of agent \(i\), \(W_p^{(\ell)}\) is a learnable projection matrix.
The group-level update aggregates over hyperedges:
\begin{equation}
    q_i^{(\ell)}
    =
    \sigma\!\left(
    \sum_{e\in\mathcal{H}(i)}
    \frac{1}{d_i}
    \frac{1}{|e|}
    \sum_{j\in \mathcal{V}(e)}
    W_q^{(\ell)} z_j^{(\ell)}
    \right),
\end{equation}
where \(\mathcal{H}(i)\) is the set of hyperedges containing agent \(i\), 
\(d_i = |\mathcal{H}(i)|\) is the hyperedge degree of agent \(i\), and 
\(|e|\) is the number of agents in hyperedge \(e\). 
This is equivalent to an unweighted degree-normalized hypergraph convolution using the node--hyperedge incidence matrix \(H\).
The operation first aggregates node representations into group representations and then broadcasts the group information back to member agents. 
Here, $W_q^{(\ell)}$ is a learnable projection matrix.
The two updates are then fused by concatenation:
\begin{equation}
    z_i^{(\ell+1)}
    =
    \phi\left(
    p_i^{(\ell)} \Vert q_i^{(\ell)}
    \right),
\end{equation}
where \(\phi\) denotes dropout applied after concatenation.
% After two layers, we apply mean and max pooling over the final node
% representations, giving
% \begin{equation}
%     h^T, h^0 \in \mathbb{R}^{256}.
% \end{equation}
Because the same encoder is used for both graphs, \(h^T\) and \(h^0\)
lie in the same representation space. We define the communication-induced shift and the final calibration representation as
\begin{equation}
    \Delta h_x^T = h^T - h^0, 
    \qquad 
    z_x = [h^T \Vert h^0 \Vert \Delta h_x^T \Vert b_x].
\end{equation}
Here \(b_x\) is a benchmark one-hot vector.
% Since $h^T$ and $h^0$ are both 256-dimensional and $b_x$ has five dimensions, $z_x$ has dimension
% \[
% 256 \times 3 + 5 = 773.
% \]
The three graph terms serve different roles. 
The observed representation $h^T$ describes the panel that produced the final answer. 
The counterfactual representation $h^0$ describes the background dependence of the same agents when they answer independently. 
The difference $h^T-h^0$ captures the topology-induced shift from independent answering to communication.

% \subsection{Graph-Centric Calibration Learning}
\subsection{Counterfactual Shift Calibration Learning}
% \textsc{CAGE-CAL} predicts a panel-level correctness probability
We train with a calibration-aware objective combining smoothed binary
cross-entropy and a Brier penalty: 

\begin{equation}
\small
\mathcal{L}
=
\mathrm{BCE}\!\left(\hat{p}(x),\widetilde{C}(x)\right)
+
\lambda_{\mathrm{Brier}}
\left(\hat{p}(x)-C(x)\right)^2
\end{equation}
where $\hat{p}(x)=\sigma(\mathrm{MLP}(z_x))$ is the predicted probability that the panel answer is correct.
The binary target is $C(x)=\mathbf{1}[\hat{y}(x)=y^\star(x)]$, where $\hat{y}(x)$ denotes the panel prediction and $y^\star(x)$ denotes the gold answer.
To regularize the binary objective, we use the smoothed target
$\widetilde{C}(x)=C(x)(1-2\alpha)+\alpha$.
The BCE term encourages discrimination between correct and incorrect panel predictions, while the Brier term directly penalizes miscalibrated probability estimates.

\section{Experiments}
\label{sec:experiments}
\begin{table*}[!t]
\centering
\setlength{\tabcolsep}{3pt}
\resizebox{\textwidth}{!}{%
\begin{tabular}{l c c c c c c c c c c c c}
\toprule
& \multicolumn{2}{c}{\textbf{TriviaQA}}
& \multicolumn{2}{c}{\textbf{TruthfulQA}}
& \multicolumn{2}{c}{\textbf{MMLU-Pro}}
& \multicolumn{2}{c}{\textbf{GSM8K}}
& \multicolumn{2}{c}{\textbf{BBH}}
& \multicolumn{2}{c}{\textbf{Mean}} \\
\cmidrule(lr){2-3}\cmidrule(lr){4-5}\cmidrule(lr){6-7}\cmidrule(lr){8-9}\cmidrule(lr){10-11}\cmidrule(lr){12-13}
\textbf{Method}
  & ECE$\downarrow$ & AUROC$\uparrow$
  & ECE$\downarrow$ & AUROC$\uparrow$
  & ECE$\downarrow$ & AUROC$\uparrow$
  & ECE$\downarrow$ & AUROC$\uparrow$
  & ECE$\downarrow$ & AUROC$\uparrow$
  & ECE$\downarrow$ & AUROC$\uparrow$ \\
\midrule
\rowcolor{gray!20}
\multicolumn{13}{c}{\textit{Post-hoc plurality calibrators}}\\
Plurality share     & 11.43 {\scriptsize $\pm$ 0.07} & 81.89 {\scriptsize $\pm$ 0.79} & 25.23 {\scriptsize $\pm$ 0.89} & 71.16 {\scriptsize $\pm$ 3.89} & 11.70 {\scriptsize $\pm$ 0.89} & 61.78 {\scriptsize $\pm$ 0.80} & 17.98 {\scriptsize $\pm$ 0.22} & 83.01 {\scriptsize $\pm$ 2.22} & 23.87 {\scriptsize $\pm$ 1.36} & 66.15 {\scriptsize $\pm$ 1.49} & 18.04 {\scriptsize $\pm$ 0.21} & 72.80 {\scriptsize $\pm$ 1.52} \\
\;+\;Platt          & 7.96 {\scriptsize $\pm$ 1.64} & 81.89 {\scriptsize $\pm$ 0.79} & 9.82 {\scriptsize $\pm$ 0.15} & 71.16 {\scriptsize $\pm$ 3.89} & 9.96 {\scriptsize $\pm$ 0.67} & 61.78 {\scriptsize $\pm$ 0.80} & 2.86 {\scriptsize $\pm$ 0.23} & 83.01 {\scriptsize $\pm$ 2.22} & 10.20 {\scriptsize $\pm$ 0.90} & 66.15 {\scriptsize $\pm$ 1.49} & 8.16 {\scriptsize $\pm$ 0.36} & 72.80 {\scriptsize $\pm$ 1.52} \\
\;+\;Isotonic       & 8.11 {\scriptsize $\pm$ 1.45} & 81.90 {\scriptsize $\pm$ 0.73} & 8.47 {\scriptsize $\pm$ 0.00} & 70.45 {\scriptsize $\pm$ 3.99} & 10.83 {\scriptsize $\pm$ 1.51} & 61.00 {\scriptsize $\pm$ 0.47} & 3.22 {\scriptsize $\pm$ 0.60} & 80.00 {\scriptsize $\pm$ 2.87} & 7.96 {\scriptsize $\pm$ 0.09} & 65.37 {\scriptsize $\pm$ 2.15} & 7.72 {\scriptsize $\pm$ 0.73} & 71.74 {\scriptsize $\pm$ 1.75} \\
\;+\;Scaling-bin.   & 9.58 {\scriptsize $\pm$ 0.19} & 81.30 {\scriptsize $\pm$ 0.23} & 8.41 {\scriptsize $\pm$ 1.88} & 70.83 {\scriptsize $\pm$ 3.62} & \textbf{7.79 {\scriptsize $\pm$ 1.04}} & 61.97 {\scriptsize $\pm$ 0.86} & \underline{2.28 {\scriptsize $\pm$ 0.34}} & 82.85 {\scriptsize $\pm$ 1.43} & \underline{6.86 {\scriptsize $\pm$ 2.42}} & 65.95 {\scriptsize $\pm$ 1.29} & \underline{6.98 {\scriptsize $\pm$ 0.07}} & 72.58 {\scriptsize $\pm$ 1.39} \\
\midrule
\rowcolor{gray!20}
\multicolumn{13}{c}{\textit{LLM-elicited confidence estimators}}\\
LLM-Cal (no topo)  & 7.99 {\scriptsize $\pm$ 0.36} & 85.74 {\scriptsize $\pm$ 1.17} & 54.88 {\scriptsize $\pm$ 0.83} & \underline{78.50 {\scriptsize $\pm$ 1.54}} & 33.19 {\scriptsize $\pm$ 2.36} & 57.64 {\scriptsize $\pm$ 1.22} & 8.96 {\scriptsize $\pm$ 0.32} & 67.18 {\scriptsize $\pm$ 1.09} & 13.87 {\scriptsize $\pm$ 0.85} & 63.92 {\scriptsize $\pm$ 0.12} & 23.78 {\scriptsize $\pm$ 0.33} & 70.60 {\scriptsize $\pm$ 0.49} \\
LLM-Cal (+topo)    & 8.71 {\scriptsize $\pm$ 0.07} & 85.73 {\scriptsize $\pm$ 0.01} & 54.52 {\scriptsize $\pm$ 0.87} & 78.48 {\scriptsize $\pm$ 2.25} & 32.03 {\scriptsize $\pm$ 1.00} & 58.33 {\scriptsize $\pm$ 0.62} & 8.28 {\scriptsize $\pm$ 0.57} & 71.16 {\scriptsize $\pm$ 5.16} & 14.40 {\scriptsize $\pm$ 1.89} & 62.56 {\scriptsize $\pm$ 0.80} & 23.59 {\scriptsize $\pm$ 0.11} & 71.25 {\scriptsize $\pm$ 1.52} \\
Collab.~Cal.       & 8.84 {\scriptsize $\pm$ 0.73} & \textbf{86.30 {\scriptsize $\pm$ 0.45}} & 52.87 {\scriptsize $\pm$ 0.95} & 78.24 {\scriptsize $\pm$ 1.16} & 31.30 {\scriptsize $\pm$ 1.84} & 59.25 {\scriptsize $\pm$ 2.35} & 8.01 {\scriptsize $\pm$ 1.50} & 70.01 {\scriptsize $\pm$ 2.72} & 11.59 {\scriptsize $\pm$ 0.08} & 64.71 {\scriptsize $\pm$ 0.48} & 22.52 {\scriptsize $\pm$ 0.13} & 71.70 {\scriptsize $\pm$ 0.30} \\
\midrule
\rowcolor{gray!20}
\multicolumn{13}{c}{\textit{Trained calibrators}}\\
Scalar~+~GBT       & 9.13 {\scriptsize $\pm$ 1.05} & 79.24 {\scriptsize $\pm$ 0.78} & 10.28 {\scriptsize $\pm$ 0.81} & 66.81 {\scriptsize $\pm$ 2.30} & 18.38 {\scriptsize $\pm$ 0.76} & 63.71 {\scriptsize $\pm$ 2.04} & 4.34 {\scriptsize $\pm$ 0.27} & 76.35 {\scriptsize $\pm$ 1.86} & 10.89 {\scriptsize $\pm$ 1.63} & 69.68 {\scriptsize $\pm$ 0.33} & 10.60 {\scriptsize $\pm$ 0.18} & 71.16 {\scriptsize $\pm$ 0.28} \\
GraphCal           & 11.05 {\scriptsize $\pm$ 0.48} & 80.06 {\scriptsize $\pm$ 1.00} & 33.66 {\scriptsize $\pm$ 1.18} & 63.75 {\scriptsize $\pm$ 3.20} & 13.33 {\scriptsize $\pm$ 0.81} & 55.18 {\scriptsize $\pm$ 0.47} & 19.70 {\scriptsize $\pm$ 0.32} & 81.02 {\scriptsize $\pm$ 1.12} & 19.77 {\scriptsize $\pm$ 0.55} & \underline{71.53 {\scriptsize $\pm$ 0.15}} & 19.50 {\scriptsize $\pm$ 0.20} & 70.31 {\scriptsize $\pm$ 0.28} \\
DiscoUQ-LLM        & 6.88 {\scriptsize $\pm$ 1.59} & 82.19 {\scriptsize $\pm$ 0.11} & \underline{6.99 {\scriptsize $\pm$ 1.17}} & 71.93 {\scriptsize $\pm$ 4.48} & 10.55 {\scriptsize $\pm$ 0.66} & \underline{64.05 {\scriptsize $\pm$ 0.76}} & 2.43 {\scriptsize $\pm$ 0.70} & \underline{83.19 {\scriptsize $\pm$ 3.28}} & 8.54 {\scriptsize $\pm$ 1.03} & 65.93 {\scriptsize $\pm$ 1.27} & 7.08 {\scriptsize $\pm$ 0.49} & \underline{73.46 {\scriptsize $\pm$ 1.94}} \\
\midrule
\rowcolor[RGB]{222,230,241}
\textbf{\methodname{} (ours)}
                   & \textbf{4.25 {\scriptsize $\pm$ 1.49}}  & \underline{86.12 {\scriptsize $\pm$ 0.78}}
                   & \textbf{6.50 {\scriptsize $\pm$ 0.72}}  & \textbf{79.66 {\scriptsize $\pm$ 2.43}}
                   & \underline{9.30 {\scriptsize $\pm$ 1.62}} & \textbf{77.74 {\scriptsize $\pm$ 0.75}}
                   & \textbf{1.64 {\scriptsize $\pm$ 0.05}}  & \textbf{84.04 {\scriptsize $\pm$ 3.55}}
                   & \textbf{6.12 {\scriptsize $\pm$ 2.57}}  & \textbf{90.48 {\scriptsize $\pm$ 0.81}}
                   & \textbf{5.56 {\scriptsize $\pm$ 0.03}}  & \textbf{83.61 {\scriptsize $\pm$ 1.34}}$^\dagger$ \\
% \midrule
% \multicolumn{13}{l}{\textit{Upper bound (uses test labels)}}\\
% Oracle isotonic    & 0.00 {\scriptsize $\pm$ 0.00} & 83.64 {\scriptsize $\pm$ 0.89} & 0.00 {\scriptsize $\pm$ 0.00} & 73.79 {\scriptsize $\pm$ 3.48} & 0.00 {\scriptsize $\pm$ 0.00} & 64.45 {\scriptsize $\pm$ 0.65} & 0.00 {\scriptsize $\pm$ 0.00} & 87.70 {\scriptsize $\pm$ 1.76} & 0.00 {\scriptsize $\pm$ 0.00} & 69.29 {\scriptsize $\pm$ 0.91} & 0.00 {\scriptsize $\pm$ 0.00} & 75.77 {\scriptsize $\pm$ 1.18} \\
\bottomrule
\end{tabular}%
}
\vspace{-5pt}
\caption{\textbf{In-distribution headline calibration results.}
ECE ($\downarrow$) and AUROC ($\uparrow$), mean $\pm$ std over 3
rollout seeds, averaged across the five topologies. Mean column
macro-averages benchmarks. \textbf{Bold} marks the best result in each column, and \underline{underline} marks the second best. $\dagger$: Wilcoxon $p{<}0.05$ vs
DiscoUQ-LLM on Mean AUROC.
Heuristic UQ baselines and AUARC / Brier for all methods are
deferred to Table~\ref{tab:heuristic-uq} and Table~\ref{tab:brier}.}
\label{tab:main}
\vspace{-15pt}
\end{table*}

% We evaluate CAGE-CAL as a panel-level confidence estimator. 
% Given a completed multi-agent panel and its plurality answer, each method estimates the confidence that the panel answer is correct.

\subsection{Experimental Setup}

We use the 25-cell grid from Section~\ref{sec:setup}, formed by five benchmarks and five communication topologies. 
The benchmarks are TriviaQA~\citep{joshi2017triviaqalargescaledistantly}, TruthfulQA~\citep{lin2022truthfulqameasuringmodelsmimic}, MMLU-Pro~\citep{wang2024mmluprorobustchallengingmultitask}, GSM8K~\citep{cobbe2021trainingverifierssolvemath}, and BIG-Bench Hard~\citep{suzgun2022challengingbigbenchtaskschainofthought}. 
The topologies are iid, debate, chain, hub-spoke, and tree. 
Each topology and benchmark cell is evaluated with three rollouts. 
For each query, the panel first produces a plurality answer, and the calibrator then scores the reliability of that answer.
Metrics are averaged over topologies and rollouts within each benchmark. 
The Mean column macro-averages the five benchmarks. 
For \select{}, we evaluate only matched test groups where all five topology outputs are available for the same query. Full experimental details are provided in Appendix~\ref{app:experimental_setup}.

\subsection{Baselines and Metrics}
\label{sec:baselines}
% such as Platt scaling~\cite{platt1999probabilistic}, isotonic regression~\cite{zadrozny2002transforming}  scaling-binning~\cite{kumar2019verified} 

We compare \methodname{} against three categories of baselines to address different evaluation dimensions: 
\textit{Post hoc plurality calibrators}:
% Plurality share uses the fraction of agents supporting the plurality answer as confidence. 
Calibrate the confidence score computed from plurality share~\cite{kuncheva2004combining} on the validation split using three  post-hoc methods: (1) Platt scaling~\cite{platt1999probabilistic}, (2) Isotonic regression~\cite{zadrozny2002transforming}, and (3) Scaling-binning~\cite{kumar2019verified};
\textit{LLM-elicited confidence estimators.} 
These baselines use an LLM judge to elicit a probability-like confidence score for the panel prediction.
They test whether panel reliability can be inferred from the final agent responses, optionally augmented with the topology description.
The baselines include: (1) LLM-Cal without topology information, (2) LLM-Cal with topology information, and (3) Collaborative Calibration~\citep{yang2024confidence};
\textit{Trained calibrators.} 
(1) Scalar + GBT~\cite{ke2017lightgbm} uses vote, confidence, and graph-summary features without relational encoding; 
(2) GraphCal~\citep{li2025graph} adapts graph-based calibration to the panel setting; 
and (3) DiscoUQ-LLM~\citep{jiang2026discouq} serves as a strong baseline based on disagreement features.
% \paragraph{Ranking-only UQ scores.}
We also evaluate answer entropy~\citep{kuhn2023semanticuncertaintylinguisticinvariances}, average log probability~\cite{kadavath2022languagemodelsmostlyknow}, DiverseAgentEntropy~\citep{feng2025rethinkingllmuncertaintymultiagent}, and MATU~\citep{chen2026responsecountsquantifyinguncertainty}. 
These methods provide uncertainty scores rather than calibrated probabilities, so we report AUROC and AUARC without interpreting the scores probabilistically.

\paragraph{Metrics.}
Our primary metrics are ECE and AUROC. 
ECE measures whether predicted confidence matches empirical correctness, while AUROC measures whether correct panel answers receive higher scores than incorrect ones. 
We additionally report Brier score and AUARC as complementary metrics. 
Brier score evaluates probability quality under a proper scoring rule, and AUARC evaluates selective prediction when panel answers are accepted in decreasing confidence order. 
For ranking-only UQ scores, we omit ECE unless a probability scale is introduced through the shared validation-set calibration protocol.

\begin{figure}[!t]
  \centering
  \includegraphics[width=\columnwidth]{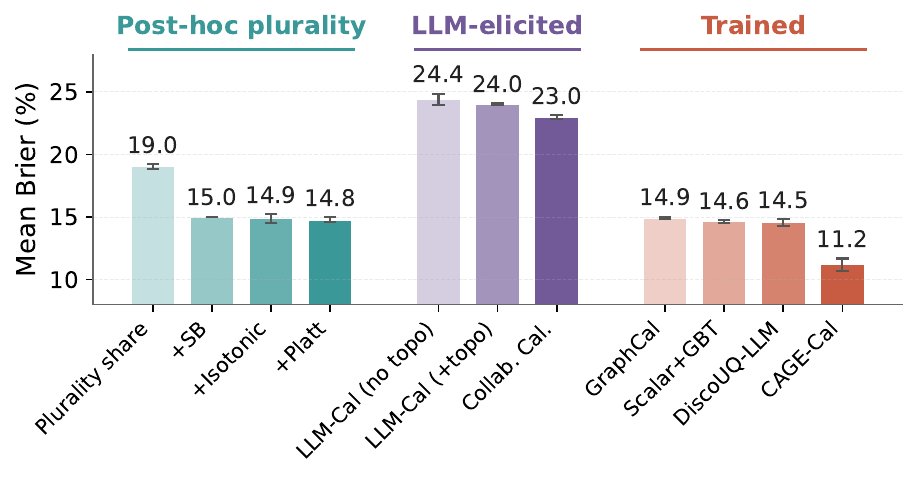}
  \vspace{-20pt}
  \caption{\textbf{Mean Brier score by method family} (lower is
  better). Within each family, bars are sorted worst $\to$ best
  (light $\to$ dark). \methodname{} (rightmost) has the lowest
  Brier overall.}
  \label{fig:brier-bars}
  \vspace{-15pt}
\end{figure}
\subsection{Main Calibration Results}
\label{sec:main_results}

Table~\ref{tab:main} reports the headline in-distribution results. 
The first block shows that post-hoc calibration can substantially reduce the ECE of plurality share, from $18.04$ to $6.98$ with scaling-binning. 
However, these methods still rely on the same scalar agreement signal. 
Platt scaling is monotonic and therefore preserves the AUROC of plurality share at $72.80$; isotonic regression and scaling-binning only change ranking slightly through ties and bins. 
This confirms that post-hoc calibration can improve probability scale, but cannot add the structural information needed to distinguish independent agreement from communication-induced consensus.
LLM-elicited confidence estimators are also limited in this setting. 
Their mean ECE remains high, ranging from $22.52$ to $23.78$, and adding topology information to the prompt brings only a small AUROC change. 
This suggests that a topology label alone is not enough for a judge model to recover the query-specific dependence among agents.
Among trained calibrators, \methodname{} achieves the best mean ECE and the strongest mean AUROC. 
Compared with DiscoUQ-LLM, the strongest prior trained baseline, \methodname{} reduces mean ECE from $7.08$ to $5.56$ and improves mean AUROC from $73.46$ to $83.61$. 
The AUROC gain is especially large on MMLU-Pro and BBH, where correlated agreement is more harmful: \methodname{} improves AUROC by $13.69$ points on MMLU-Pro and $24.55$ points on BBH over DiscoUQ-LLM. 
These gains support the central claim that panel confidence should depend not only on how many agents agree, but also on how that agreement was formed.
Figure~\ref{fig:brier-bars} provides a complementary view through Brier score. 
\methodname{} achieves the lowest mean Brier score at $11.2$, outperforming DiscoUQ-LLM ($14.5$), Scalar + GBT ($15.6$), and GraphCal ($19.2$). 
Thus, the improvement is not only an artifact of ECE binning; \methodname{} also produces sharper and better scaled probability estimates.

\subsection{Ranking-Only UQ Comparison}
\label{sec:uq_ranking}
\begin{figure}[!t]
  \centering
  \includegraphics[width=\columnwidth]{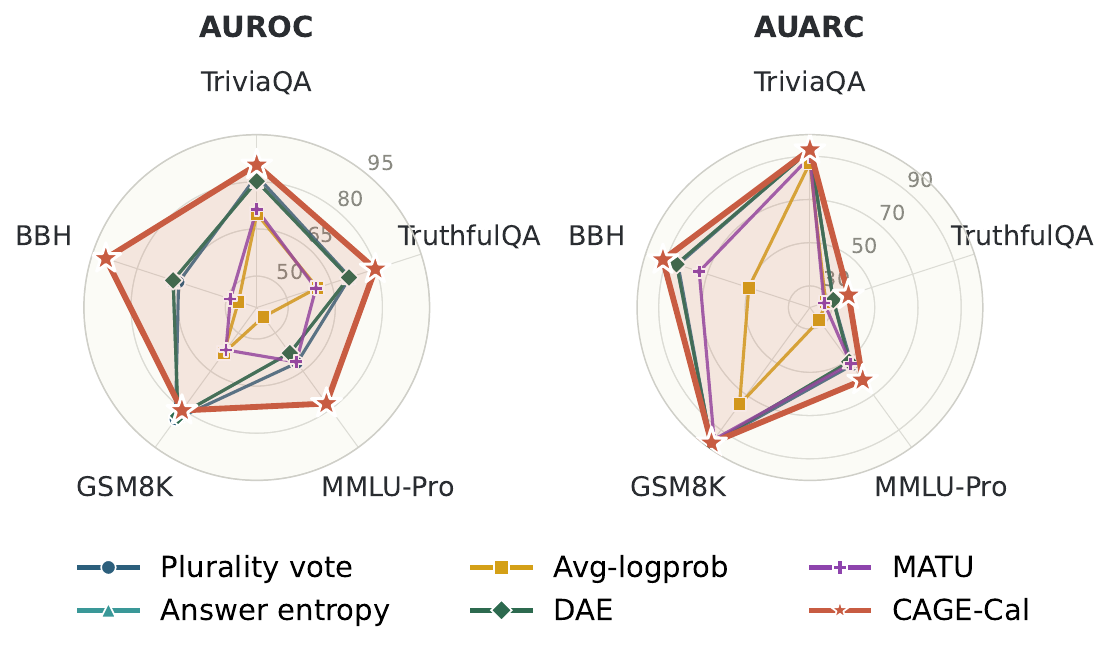}
  \vspace{-20pt}
  \caption{\textbf{AUROC and AUARC of \methodname{} vs.\ heuristic
  UQ baselines.} Each axis reports benchmark-level mean over 3 rollouts and 5 topologies. Per-benchmark numbers in
  Appendix Table~\ref{tab:heuristic-uq}.}
  \label{fig:auroc-auarc}
  \vspace{-15pt}
\end{figure}

Figure~\ref{fig:auroc-auarc} compares \methodname{} with ranking-only UQ baselines using AUROC and AUARC. 
These methods provide useful uncertainty scores, but they do not define a calibrated probability scale. 
Plurality vote, entropy, average log probability, DiverseAgentEntropy, and MATU all summarize the panel with scalar signals. 
Such signals work when disagreement reflects uncertainty, but they miss the difference between benign diversity and false consensus.
\methodname{} gives the strongest overall reliability ranking across the five benchmarks. 
The advantage is most visible on BBH and MMLU-Pro, where scalar disagreement signals are less reliable. 
This aligns with the failure-mode analysis: when agents become correlated through shared model families or communication paths, the reliability of the final answer depends on the dependency structure behind the vote, not only on the vote distribution itself.

\subsection{Confidence-Routed Topology Selection}
\label{sec:cage_select}

\begin{figure}[!t]
  \centering
  \includegraphics[width=0.9\columnwidth]{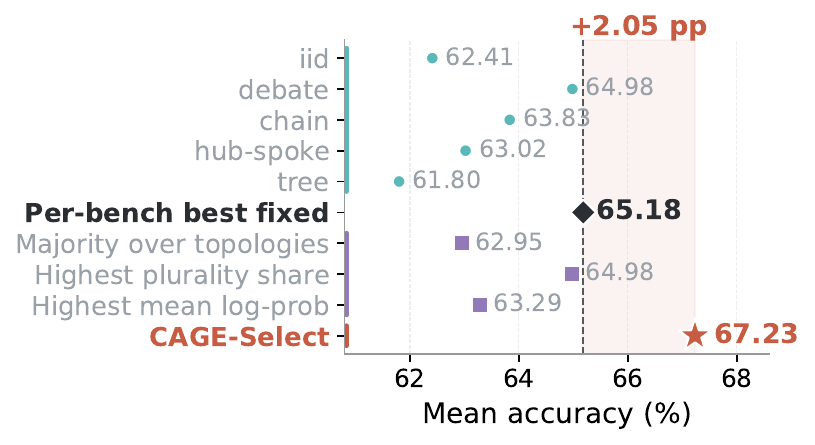}
  \vspace{-10pt}
  \caption{\textbf{Mean accuracy of routing strategies.}
  Dashed line marks per-bench best fixed ($65.18$). Per-bench
  breakdown in Appendix Table~\ref{tab:cage-select}.}
  \label{fig:cage-select}
  \vspace{-15pt}
\end{figure}

We further test whether calibrated confidence can serve as a control signal for multi-agent inference. 
Instead of committing to a single topology for all queries, \select{} uses \methodname{} confidence to choose which topology output should be trusted for each query.
Figure~\ref{fig:cage-select} shows that the best fixed topology reaches $65.18\%$ mean accuracy, while simple routing rules based on plurality share or mean log probability do not improve over it. 
\select{} reaches $67.23\%$, a $+2.05$ point gain. 
Thus, \methodname{} confidence is not only calibrated within a topology, but also comparable across topologies, making it useful as a control signal for multi-agent inference.

\section{Analysis}
\label{sec:analysis}

\begin{table*}[!t]
\centering
\setlength{\tabcolsep}{4pt}
\resizebox{\textwidth}{!}{%
\begin{tabular}{l c c c c c c c c c c c c}
\toprule
& \multicolumn{2}{c}{\textbf{TriviaQA}}
& \multicolumn{2}{c}{\textbf{TruthfulQA}}
& \multicolumn{2}{c}{\textbf{MMLU-Pro}}
& \multicolumn{2}{c}{\textbf{GSM8K}}
& \multicolumn{2}{c}{\textbf{BBH}}
& \multicolumn{2}{c}{\textbf{Mean}} \\
\cmidrule(lr){2-3}\cmidrule(lr){4-5}\cmidrule(lr){6-7}\cmidrule(lr){8-9}\cmidrule(lr){10-11}\cmidrule(lr){12-13}
\textbf{Method}
  & ECE$\downarrow$ & AUROC$\uparrow$
  & ECE$\downarrow$ & AUROC$\uparrow$
  & ECE$\downarrow$ & AUROC$\uparrow$
  & ECE$\downarrow$ & AUROC$\uparrow$
  & ECE$\downarrow$ & AUROC$\uparrow$
  & ECE$\downarrow$ & AUROC$\uparrow$ \\
\midrule
Scalar~+~GBT       & 11.15 {\scriptsize $\pm$ 4.43} & 80.28 {\scriptsize $\pm$ 4.95} & 23.50 {\scriptsize $\pm$ 12.68} & 70.46 {\scriptsize $\pm$ 2.14} & 16.39 {\scriptsize $\pm$ 4.04} & 60.73 {\scriptsize $\pm$ 4.28} & 10.14 {\scriptsize $\pm$ 3.84} & 80.58 {\scriptsize $\pm$ 2.49} & 13.33 {\scriptsize $\pm$ 4.22} & 62.48 {\scriptsize $\pm$ 2.54} & 14.90 {\scriptsize $\pm$ 1.73} & 70.91 {\scriptsize $\pm$ 2.54} \\
GraphCal           & 11.93 {\scriptsize $\pm$ 4.66} & 80.13 {\scriptsize $\pm$ 5.01} & 34.15 {\scriptsize $\pm$ 9.54} & 63.51 {\scriptsize $\pm$ 2.89} & 14.41 {\scriptsize $\pm$ 2.17} & 54.79 {\scriptsize $\pm$ 5.86} & 19.59 {\scriptsize $\pm$ 6.43} & 80.92 {\scriptsize $\pm$ 6.54} & 20.05 {\scriptsize $\pm$ 4.09} & 70.62 {\scriptsize $\pm$ 3.60} & 20.03 {\scriptsize $\pm$ 1.87} & 70.00 {\scriptsize $\pm$ 2.40} \\
DiscoUQ-LLM       & 12.52 {\scriptsize $\pm$ 5.24} & 82.60 {\scriptsize $\pm$ 4.90} & 24.42 {\scriptsize $\pm$ 13.93} & 71.78 {\scriptsize $\pm$ 3.72} & 16.83 {\scriptsize $\pm$ 4.67} & 59.37 {\scriptsize $\pm$ 7.95} & 10.09 {\scriptsize $\pm$ 4.12} & \textbf{84.59 {\scriptsize $\pm$ 3.95}} & 15.25 {\scriptsize $\pm$ 4.40} & 63.00 {\scriptsize $\pm$ 3.58} & 15.82 {\scriptsize $\pm$ 8.45} & 72.27 {\scriptsize $\pm$ 11.32} \\
\rowcolor[RGB]{222,230,241}
\textbf{\methodname{} (ours)}
                  & \textbf{4.63 {\scriptsize $\pm$ 0.71}}  & \textbf{85.74 {\scriptsize $\pm$ 1.41}}
                  & \textbf{8.70 {\scriptsize $\pm$ 1.84}}  & \textbf{79.59 {\scriptsize $\pm$ 4.67}}
                  & \textbf{11.19 {\scriptsize $\pm$ 1.29}} & \textbf{76.67 {\scriptsize $\pm$ 3.63}}
                  & \textbf{1.89 {\scriptsize $\pm$ 1.31}}  & 80.39 {\scriptsize $\pm$ 5.75}
                  & \textbf{6.38 {\scriptsize $\pm$ 1.38}}  & \textbf{88.67 {\scriptsize $\pm$ 1.12}}
                  & \textbf{6.56 {\scriptsize $\pm$ 0.68}}  & \textbf{82.21 {\scriptsize $\pm$ 1.84}} \\
\bottomrule
\end{tabular}%
}
\vspace{-5pt}
\caption{\textbf{Leave-one-topology-out (LOTO) generalization.}
ECE ($\downarrow$) and AUROC ($\uparrow$), mean over the 5
held-out-topology folds. Trained calibrators only.}
\label{tab:main-loto}
\vspace{-15pt}
\end{table*}

\begin{table}[!h]
\centering
\setlength{\tabcolsep}{4pt}
\resizebox{\columnwidth}{!}{%
\begin{tabular}{l c c c}
\toprule
\textbf{Variant}                          & \textbf{ECE} $\downarrow$
                                            & \textbf{AUROC} $\uparrow$
                                            & \textbf{AUARC} $\uparrow$ \\
\midrule
\rowcolor{gray!20}
\multicolumn{4}{c}{\textit{Scalar-summary baselines}}\\
Scalar summaries + LR                & $8.03${\scriptsize$\pm0.35$}
                                             & $74.12${\scriptsize$\pm1.24$}
                                             & $71.64${\scriptsize$\pm0.69$} \\
Scalar summaries + GBT                & $7.99${\scriptsize$\pm0.95$}
                                             & $74.78${\scriptsize$\pm1.32$}
                                             & $72.70${\scriptsize$\pm0.50$} \\
\midrule
\rowcolor{gray!20}
\multicolumn{4}{c}{\textsc{CAGE-Cal} incremental variants}\\
Observed graph encoder only               & $6.97${\scriptsize$\pm0.28$}
                                             & $81.25${\scriptsize$\pm0.88$}
                                             & $75.61${\scriptsize$\pm0.13$} \\
\,\,+ Iid counterfactual tower                  & $6.78${\scriptsize$\pm0.33$}
                                             & $81.89${\scriptsize$\pm1.47$}
                                             & $75.79${\scriptsize$\pm0.24$} \\
% \rowcolor[RGB]{222,230,241}
% \multicolumn{1}{c}{$\Delta$ vs. previous}
% & {$-0.19$}
% & {$+0.64$}
% & {$+0.18$} \\
\,\,+ Group-level hyperedge stream                      & $6.75${\scriptsize$\pm0.37$}
                                             & $82.56${\scriptsize$\pm1.27$}
                                             & $76.08${\scriptsize$\pm0.38$} \\
% \rowcolor[RGB]{222,230,241}
% \multicolumn{1}{c}{$\Delta$ vs. previous}
% & {$-0.03$}
% & {$+0.67$}
% & {$+0.29$} \\

\,\,+ Calibration-aware objective (full)         & \textbf{5.56{\scriptsize$\pm0.03$}}
                                             & \textbf{83.61{\scriptsize$\pm1.34$}}
                                             & \textbf{76.47{\scriptsize$\pm0.37$}} \\
\rowcolor[RGB]{222,230,241}
\multicolumn{1}{c}{$\Delta$ vs. base}
& {$-1.41$}
& {$+2.36$}
& {$+0.86$} \\
\bottomrule
\end{tabular}%
}
\vspace{-5pt}
\caption{\textbf{Component ablation of \textsc{CAGE-CAL}}, averaged over 25 in-distribution cells 
(percent, mean $\pm$ std over 3 rollouts). 
$\Delta$ rows show absolute changes from the previous variant in percentage points. 
LR and GBT denote logistic regression and gradient-boosted trees.}
\label{tab:ablation}
\vspace{-15pt}
\end{table}

\subsection{Component Ablation}
\label{sec:ablation}

Table~\ref{tab:ablation} ablates the main components of \methodname{}. 
Scalar-summary baselines compress each panel into fixed statistics and remain far below graph-based variants, with the stronger GBT head reaching only $74.78$ AUROC and $72.70$ AUARC. 
The observed graph encoder raises AUROC to $81.25$, showing that panel reliability depends on relational structure beyond aggregate vote and confidence summaries. 
Adding the IID counterfactual tower further improves AUROC by $0.64$ points, while the group-level hyperedge stream adds another $0.67$ points by capturing shared family, role, answer-cluster, and exposure effects. 
The calibration-aware objective gives the largest ECE reduction, from $6.75$ to $5.56$, without sacrificing ranking quality. 
Overall, the gains come from modeling a counterfactual dependence shift, not from simply adding a stronger prediction head.

\subsection{Generalization to Held-Out Topologies}
\label{sec:loto}

We test topology generalization with leave-one-topology-out evaluation. 
Each fold removes one topology from both training and validation, and evaluates the calibrator on that unseen topology.
As shown in Table~\ref{tab:main-loto}, \methodname{} remains stable under this shift. 
Its mean AUROC is $82.21$, close to the in-distribution result of $83.61$, and its mean ECE rises only from $5.56$ to $6.56$. 
By comparison, DiscoUQ-LLM reaches $72.27$ mean AUROC and $15.82$ mean ECE. 
This indicates that scalar disagreement features do not transfer as well when the communication structure changes.
The result supports the relational design of \methodname{}. 
Rather than relying on a topology label, it uses communication edges, local failure correlations, answer clusters, and group-level dependency units. 
These features are defined for any topology, allowing the calibration rule to transfer to unseen communication structures.

\subsection{Correcting the Two Failure Modes}
\label{sec:failure_diagnostics}

\begin{figure}[!t]
  \centering
  \includegraphics[width=\columnwidth]{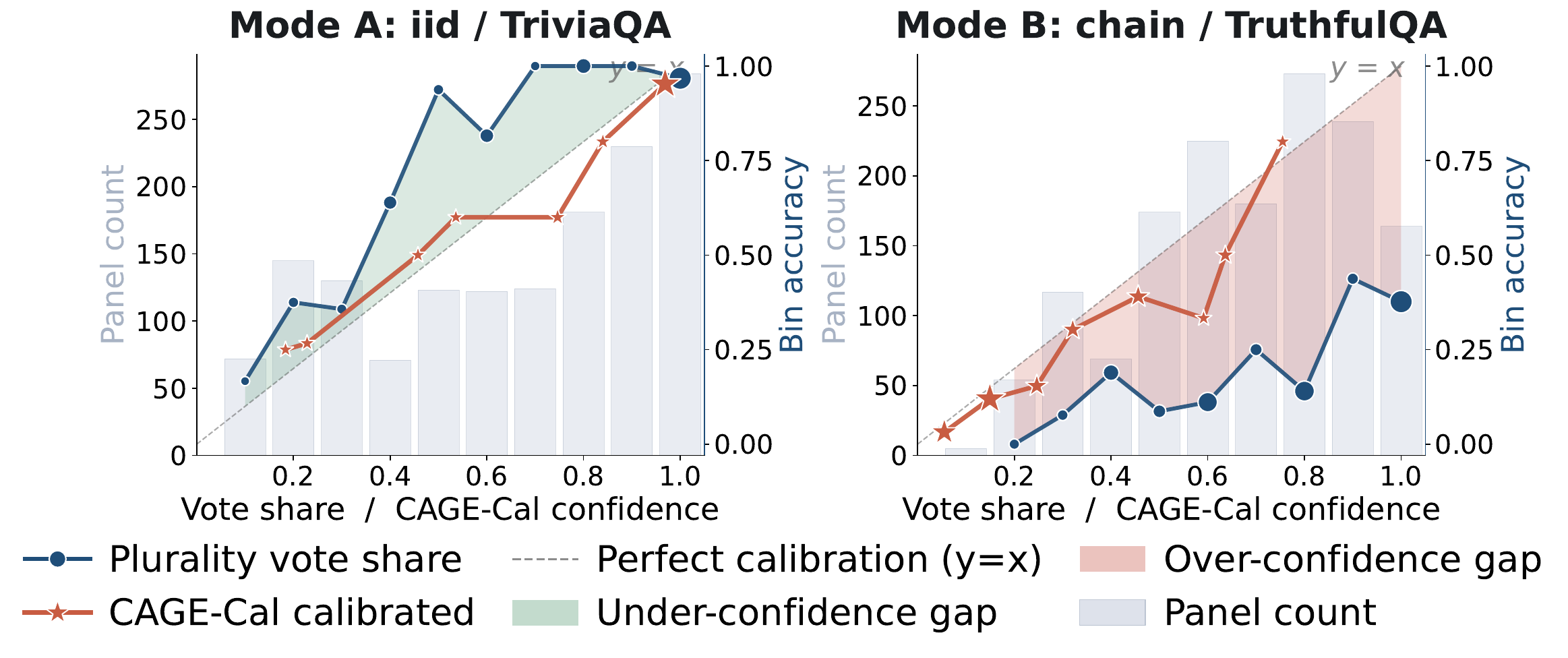}
  \vspace{-20pt}
  \caption{\textbf{Failure-mode correction.} 
Bars show panel counts, and curves show empirical bin accuracy when examples are binned by plurality share or by \methodname{} confidence. 
\methodname{} reduces under-confidence in Mode A and over-confidence in Mode B.
}
  \label{fig:failure_fix}
\vspace{-15pt}
\end{figure}

Figure~\ref{fig:failure_fix} tests whether \methodname{} corrects the two calibration failures identified earlier. 
In Mode A, iid/TriviaQA, plurality share is under-confident because wrong answers are dispersed across weak clusters. 
In Mode B, chain/TruthfulQA, plurality share is over-confident because communication can concentrate agents on the same wrong answer. 
In both cases, \methodname{} moves the reliability curve closer to the perfect calibration line. 
Thus, the same vote share can receive different confidence depending on how the agreement or disagreement was formed.
\section{Conclusion}
\label{sec:conclusion}

We investigate confidence calibration in multi-agent LLM systems and show that agreement alone is an unreliable confidence signal.
Our analysis across different benchmarks and communication topologies identifies two recurring failure modes: DUC and COC.
% Diverse agents can produce benign disagreement that leads to under-confidence, while communication can induce correlated agreement and lead to over-confidence. 
We propose \methodname{}, a counterfactual agent-graph calibration framework that contrasts post-communication graphs with IID counterfactual graphs to separate independent evidence from correlated failure, which improves reliability discrimination with competitive calibration error. 
We further introduce \select{}, which uses calibrated confidence to dynamically select the most reliable topology and improve final panel accuracy. 
Overall, our results highlight the importance of topology-aware calibration for reliable multi-agent LLM systems.
% \newpage
% \section*{Limitations}
\section{Limitations}
A limitation is that agent participation is not optimized. 
We construct panels from a predefined agent pool, while real systems may benefit from query-adaptive agent selection, where the system decides which agents are most useful for a given query.
Another limitation is that communication is not jointly optimized with calibration. 
We evaluate a fixed set of candidate communication patterns, but practical systems may need adaptive communication routing, where agents decide which interactions are most useful based on intermediate answers and uncertainty. 
Extending \methodname{} to jointly support adaptive agent selection and communication routing is an important future direction~\cite{li2026same,zhang2025agentrouter,shi2026ng,huang2026evolverouter}.
% \input{sections/ethics}

% \section*{Acknowledgments}
% \input{sections/limitations}


\bibliography{custom}

\appendix
\appendix

\section{Additional Related Work}
\label{app:related_work}

\label{con:related_work_for_graph}
\paragraph{Graph-based multi-agent design and calibration under dependence.}
%  why Graph
A related line of work represents multi-agent systems as graphs and optimizes their communication structure. 
GPTSwarm~\citep{zhuge2024languageagentsoptimizablegraphs} treats language agents as optimizable graphs, G-Designer~\citep{zhang2025gdesignerarchitectingmultiagentcommunication} learns communication topologies with graph neural networks, and AgentPrune~\citep{zhang2024cutcrapeconomicalcommunication} removes unnecessary agents or communication links for efficiency. 
These methods focus mainly on task accuracy, routing, or computational cost. 
They do not ask how a given topology changes the panel's joint failure distribution, nor do they produce calibrated confidence for the panel answer. 
Our work turns graph structure into a calibration signal: given a topology and the resulting panel outputs, we estimate whether the panel's agreement reflects independent evidence or correlated failure.

\section{Failure-Mode Analysis and Counterfactual Motivation}
\label{app:failure_analysis}
\subsection{Why Calibration Needs Counterfactual Graph Shifts}
\label{con:analysis_progress}
In this section, we first explain why confidence calibration requires both the iid graph \(G_x^0\) and the post-communication graph \(G_x^T\). We then provide empirical evidence that the shift between \(G_x^0\) and \(G_x^T\) matters for confidence calibration.

\subsubsection{Why Calibration Needs the IID and Post-Communication Graphs}
\label{con:why_two_graphs}
% \subsection{Correlated Uncertainty Graphs}
% \label{sec:correlation}

\begin{figure}[!t]
  \centering
  \includegraphics[width=\columnwidth]{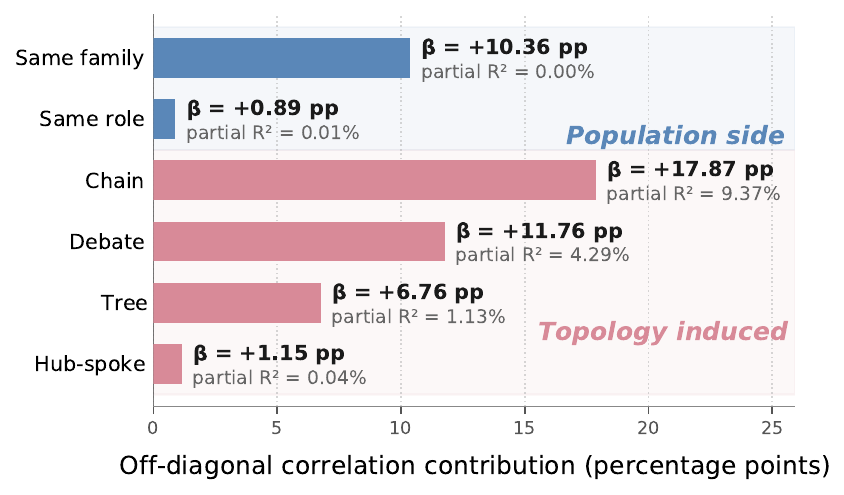}
  \caption{\textbf{Sources of agent error correlation.} OLS
  coefficients on predictors of pair correlation $\W_{ij}$.
  Communication structure dominates the backbone, which dominates
  the prompting role.}
  \label{fig:w-decomp}
\end{figure}

\begin{table}[!t]
\centering
\small
\setlength{\tabcolsep}{6pt}
\begin{tabular}{l c c}
\toprule
\textbf{Predictor} & $\bm{\beta}$ \textbf{(pp)} & \textbf{Partial $R^2$ (\%)} \\
\midrule
Intercept (baseline iid pair)        & 35.40        & ---  \\
\midrule
\multicolumn{3}{l}{\textit{Population-side sources}}\\
Same backbone family                 & +10.36       & 0.00  \\
Same prompting role                  & +0.89        & 0.01  \\
\midrule
\multicolumn{3}{l}{\textit{Topology-induced sources (vs.\ iid baseline)}}\\
\texttt{is\_chain}                   & +17.87       & 9.37  \\
\texttt{is\_debate}                  & +11.76       & 4.29  \\
\texttt{is\_tree}                    & +6.76        & 1.13  \\
\texttt{is\_hub\_spoke}              & +1.15        & 0.04  \\
\midrule
Full-model $R^2$                     & \multicolumn{2}{c}{29.02\%} \\
$n$ pairs                            & \multicolumn{2}{c}{$4{,}325$} \\
\bottomrule
\end{tabular}
\caption{\textbf{Decomposition of agent error correlation
$\W_{ij}$.} OLS on $4{,}325$ agent pairs from the 25 cells.
$\beta$ is the coefficient on the 0/1 predictor in percentage
points; partial $R^2$ is the variance lost when dropping that
predictor.}
\label{tab:w-decomp}
\end{table}

Section~\ref{sec:setup} defines $W(x)$ as the local dependency structure of a panel. 
We now ask what creates this dependency. 
If agents provided independent evidence, the off-diagonal entries of $W(x)$ would be small and weakly structured. 
Instead, we find two systematic sources of correlation: population-side similarity among agents and communication-induced coupling from the topology. 
This distinction is central to calibration. 
A topology-agnostic score can observe the final vote distribution, but it cannot tell whether the distribution was produced by independent evidence or correlated failure.

\paragraph{Where does failure correlation come from?}

For each topology and benchmark cell, we aggregate the off-diagonal entries of $W(x)$ and fit an ordinary least squares model over agent pairs:
\begin{align*}
  \W_{ij}^{(c)} = \beta_0 \;&+\; \beta_F \mathbf{1}[\text{\small{same family}}]
              \;+\; \beta_R \mathbf{1}[\text{\small{same role}}] \\
            &+\; \sum_{t \in \mathcal{T}\setminus\{\text{iid}\}}
                \beta_t \mathbf{1}[\text{\small{topology}}_{c}=t]
              \;+\; \epsilon.
\end{align*}
Figure~\ref{fig:w-decomp} visualizes the main coefficients, while the full regression results are reported in Table~\ref{tab:w-decomp}.
Table~\ref{tab:w-decomp} decomposes pairwise agent failure correlation into population-side and topology-induced sources. 
The coefficients do not measure final panel accuracy; instead, they measure how much each factor increases the tendency of two agents to succeed or fail together. 
Shared backbone families increase pairwise correlation, suggesting common model-side blind spots, while communication topologies such as chain and debate further amplify correlated failures. 
These results support our motivation for modeling both population-side dependence and topology-induced coupling in multi-agent confidence calibration.

The decomposition gives two observations. 

\begin{itemize}
\item \textbf{First, heterogeneous model families matter more than heterogeneous prompts.} 
Agents with the same backbone tend to fail together even when they use different prompting roles. 
In contrast, shared prompting role contributes little once backbone family is controlled. 
This suggests that prompt diversity can make a panel appear diverse without fully decorrelating its errors.
\item \textbf{Second, communication topology is a major source of additional dependence.}
Chain has the largest topology coefficient, followed by debate and tree. 
This order follows the amount of peer exposure in each topology: later agents in a chain see many previous outputs, debate agents see many peers in one round, and tree agents receive information through local branches. 
Hub-spoke is close to the iid baseline because spokes do not see one another. 
Thus, part of the dependence is inherited from the heterogeneous agent population, while another part is induced by communication.
\end{itemize}

% \subsection{Two sources, two graphs}
\paragraph{Motivating the Counterfactual Graph Pair}
This split directly motivates the graph pair used by \methodname{}. 
A counterfactual iid graph $G_x^0$ captures the dependence that would exist without communication. 
It represents population-side structure, including shared model families, shared biases, and shared prompting conventions. 
A post-communication graph $G_x^T$ captures the same matched agents after topology $T$ has reshaped their outputs. 
It contains both population-side dependence and communication-induced dependence.
%%%
Because $G_x^0$ and $G_x^T$ are built over the same agent identities, their contrast gives a query-specific view of how communication changes the panel. 
This is the information that scalar disagreement scores discard. 
Section~\ref{sec:method} turns this counterfactual graph pair into the input representation for \methodname{}.

\subsection{Communication-Induced Correctness Shifts}
\label{sec:shifts}

Section~\ref{con:why_two_graphs} shows that communication can increase failure correlation among agents. 
We now examine the same effect at the agent level.
Communication can change an agent's correctness in both directions. 
It can help an initially wrong agent become correct, but it can also make an initially correct agent become wrong. 
Mean accuracy only reflects the net balance of these two effects. 
Calibration depends on how these shifted votes are distributed in the final panel.

\begin{figure}[!t]
  \centering
  \includegraphics[width=\columnwidth]{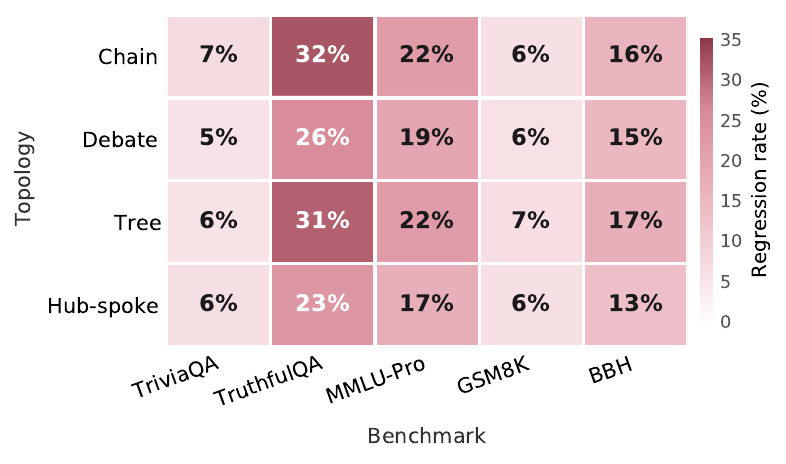}
  \caption{\textbf{Per-agent regression rate.} Fraction of
  iid-correct agents that become wrong under each topology on the
  same (question, rollout). Substantial on hard benchmarks even
  in topologies whose mean accuracy is unchanged.}
  \label{fig:regression-heatmap}
\end{figure}

\paragraph{Regression and improvement.}
For agent $i$, query $x$, rollout $r$, and topology $\tau$, we call the agent \emph{communication-regressed} when
$
c_i^{(\mathrm{iid})}(x,r)=1
\quad\text{and}\quad
c_i^{(\tau)}(x,r)=0 .
$
That is, the agent answers correctly when run independently, but answers incorrectly after communication. 
We call the symmetric case \emph{communication-improved}, when
$
c_i^{(\mathrm{iid})}(x,r)=0
\quad\text{and}\quad
c_i^{(\tau)}(x,r)=1 .
$
For each topology $\tau$ and benchmark $b$, we report both the regression rate $\mathrm{reg}_{\tau,b}$ and the improvement rate $\mathrm{impr}_{\tau,b}$. 
The iid run provides a matched counterfactual for the same agent on the same query.

\paragraph{Communication can help accuracy while hurting calibration.}
Figure~\ref{fig:regression-heatmap} shows that regression is not rare. 
On hard benchmarks, a substantial fraction of independently correct agents become wrong after communication. 
For example, chain on TruthfulQA regresses nearly one third of agents that were correct in the iid setting. 
Tree shows a similar pattern on TruthfulQA and MMLU-Pro.

At the same time, regression does not necessarily imply lower mean accuracy. 
Communication can also improve agents that were initially wrong. 
A topology can therefore be helpful for average accuracy while still injecting many wrong votes into the panel. 
This distinction is central to calibration. 
A corrected vote and a regressed vote both appear as confident votes in the final answer distribution. 
Plurality vote cannot tell whether a larger answer cluster was formed by independent correction or by harmful influence.

The key implication is that communication changes the reliability of votes, not only their average correctness. 
Two topologies can have similar mean accuracy but very different calibration behavior, because their correctness shifts land in different answer clusters. 
The next section shows how these shifts lead to two opposite calibration failures: diversity-induced under-confidence and communication-induced over-confidence.

\paragraph{Communication Can Both Correct and Corrupt Agents}\
\label{con:com_correct_ir_correct}
\begin{table}[!t]
\centering
\setlength{\tabcolsep}{4pt}
\resizebox{\columnwidth}{!}{%
\begin{tabular}{l l c c c}
\toprule
\textbf{Topology} & \textbf{Benchmark}
                  & \textbf{Reg.\ rate}
                  & \textbf{Impr.\ rate}
                  & $\bm{\Delta}$\textbf{Acc (pp)} \\
\midrule
\multirow{5}{*}{Debate}    & TriviaQA    & 5.44\%  & 39.55\% & +7.44 \\
                           & TruthfulQA  & 25.60\% & 11.05\% & +3.16 \\
                           & MMLU-Pro    & 19.15\% & 20.24\% & +4.16 \\
                           & GSM8K       & 5.86\%  & 54.65\% & +5.90 \\
                           & BBH         & 14.99\% & 28.68\% & +2.78 \\
\midrule
\multirow{5}{*}{Chain}     & TriviaQA    & 7.05\%  & 35.10\% & +5.02 \\
                           & TruthfulQA  & 31.73\% & 9.49\%  & +0.62 \\
                           & MMLU-Pro    & 21.52\% & 21.78\% & +4.10 \\
                           & GSM8K       & 6.12\%  & 54.53\% & +5.66 \\
                           & BBH         & 15.74\% & 28.92\% & +2.43 \\
\midrule
\multirow{5}{*}{Hub-spoke} & TriviaQA    & 6.31\%  & 15.86\% & +0.04 \\
                           & TruthfulQA  & 23.12\% & 6.04\%  & -0.24 \\
                           & MMLU-Pro    & 17.09\% & 12.69\% & +0.53 \\
                           & GSM8K       & 6.03\%  & 24.85\% & -0.03 \\
                           & BBH         & 13.31\% & 19.79\% & +0.16 \\
\midrule
\multirow{5}{*}{Tree}      & TriviaQA    & 5.64\%  & 31.45\% & +5.02 \\
                           & TruthfulQA  & 30.52\% & 7.51\%  & -0.26 \\
                           & MMLU-Pro    & 21.99\% & 16.97\% & +0.94 \\
                           & GSM8K       & 6.62\%  & 43.69\% & +2.64 \\
                           & BBH         & 17.19\% & 24.34\% & -0.16 \\
\bottomrule
\end{tabular}%
}
\caption{\textbf{Communication-induced regression rates.}
\emph{Reg.} $=$ fraction of iid-correct agents that became wrong
under the row's topology on the same (question, rollout).
\emph{Impr.} is the converse. $\Delta$Acc is the net per-agent
accuracy change in percentage points.}
\label{tab:regression}
\end{table}

Table~\ref{tab:regression} analyzes how communication changes individual agent correctness relative to the iid setting. 
The regression rate measures the fraction of iid-correct agents that become incorrect after communication, while the improvement rate measures the reverse transition from incorrect to correct. 
Communication can improve many agents, especially on GSM8K, but it can also corrupt initially correct agents, with particularly high regression rates on TruthfulQA. 
This shows that communication reshapes agent-level failure patterns rather than simply improving accuracy, motivating calibration methods that account for topology-induced dependence.

\subsection{Calibration Error Varies Across Communication Topologies}
\label{con:calibation_error_varios_topo}

\begin{table}[!t]
\centering
\small
\setlength{\tabcolsep}{6pt}
\begin{tabular}{l c}
\toprule
\textbf{Topology} & \textbf{Mean ECE} \\
\midrule
debate     & $14.59$ {\scriptsize $\pm$ 8.26} \\
tree       & $14.82$ {\scriptsize $\pm$ 6.09} \\
hub-spoke  & $17.55$ {\scriptsize $\pm$ 8.02} \\
iid        & $18.00$ {\scriptsize $\pm$ 8.60} \\
chain      & $19.48$ {\scriptsize $\pm$ 16.31} \\
\bottomrule
\end{tabular}
\caption{\textbf{Per-topology mean majority-vote ECE} (\%, mean
$\pm$ std across benchmarks). Chain is worst-calibrated and
benchmark-variant (benign on TriviaQA, catastrophic on
TruthfulQA).}
\label{tab:ece-by-topology}
\end{table}

Table~\ref{tab:ece-by-topology} reports the majority-vote ECE for each communication topology, averaged across benchmarks. Unlike plurality accuracy, which remains relatively stable across topologies, calibration error varies more substantially. In particular, chain has the highest mean ECE and the largest variance, suggesting that communication structure can make vote-based confidence unreliable even when final-answer accuracy changes little. This motivates our focus on calibrating panel confidence rather than only improving the predicted answer.

\subsection{Vote Share Is Not a Reliable Confidence Estimate}
\label{con:vote_share}
\begin{figure*}[!t]
  \centering
  \includegraphics[width=\textwidth]{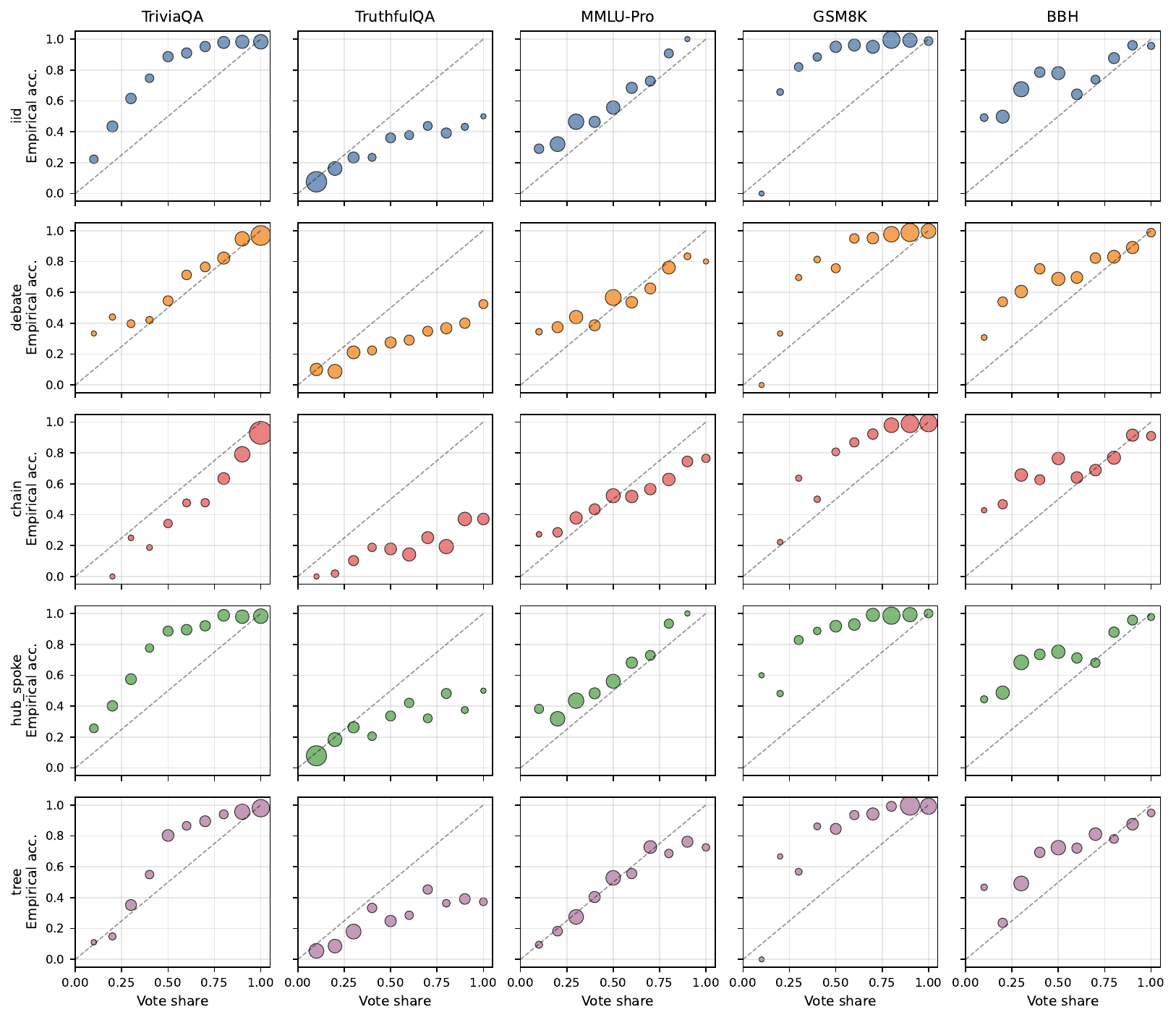}
  \caption{\textbf{Reliability diagrams per (topology, benchmark)
  cell.} Plurality vote share ($x$) vs.\ empirical accuracy
  ($y$). Bubble area is the bin's panel count. Points above the
  identity line are under-confident (Mode A); below, over-confident
  (Mode B).}
  \label{fig:reliability-per-cell}
\end{figure*}
Figure~\ref{fig:reliability-per-cell} shows that plurality vote share has different calibration behavior across benchmarks and communication topologies. 
Each point compares a vote-share bin with the empirical accuracy of panel predictions in that bin; points on the diagonal indicate perfect calibration. 
The patterns are highly cell-dependent: on TriviaQA and GSM8K, many points lie above the diagonal, meaning that vote share underestimates correctness, while on TruthfulQA many points lie below the diagonal, meaning that vote share overestimates correctness. 
Moreover, the same benchmark can behave differently under different topologies, such as iid, debate, chain, hub-spoke, and tree. 
This shows that vote share alone cannot determine panel confidence; calibration must account for both task domain and the communication structure that produced the vote.

\subsection{Agent Error Correlations Vary Across Topologies and Benchmarks}
\label{con:Agent_error_correlations}
\begin{figure*}[!t]
  \centering
  \includegraphics[width=\textwidth]{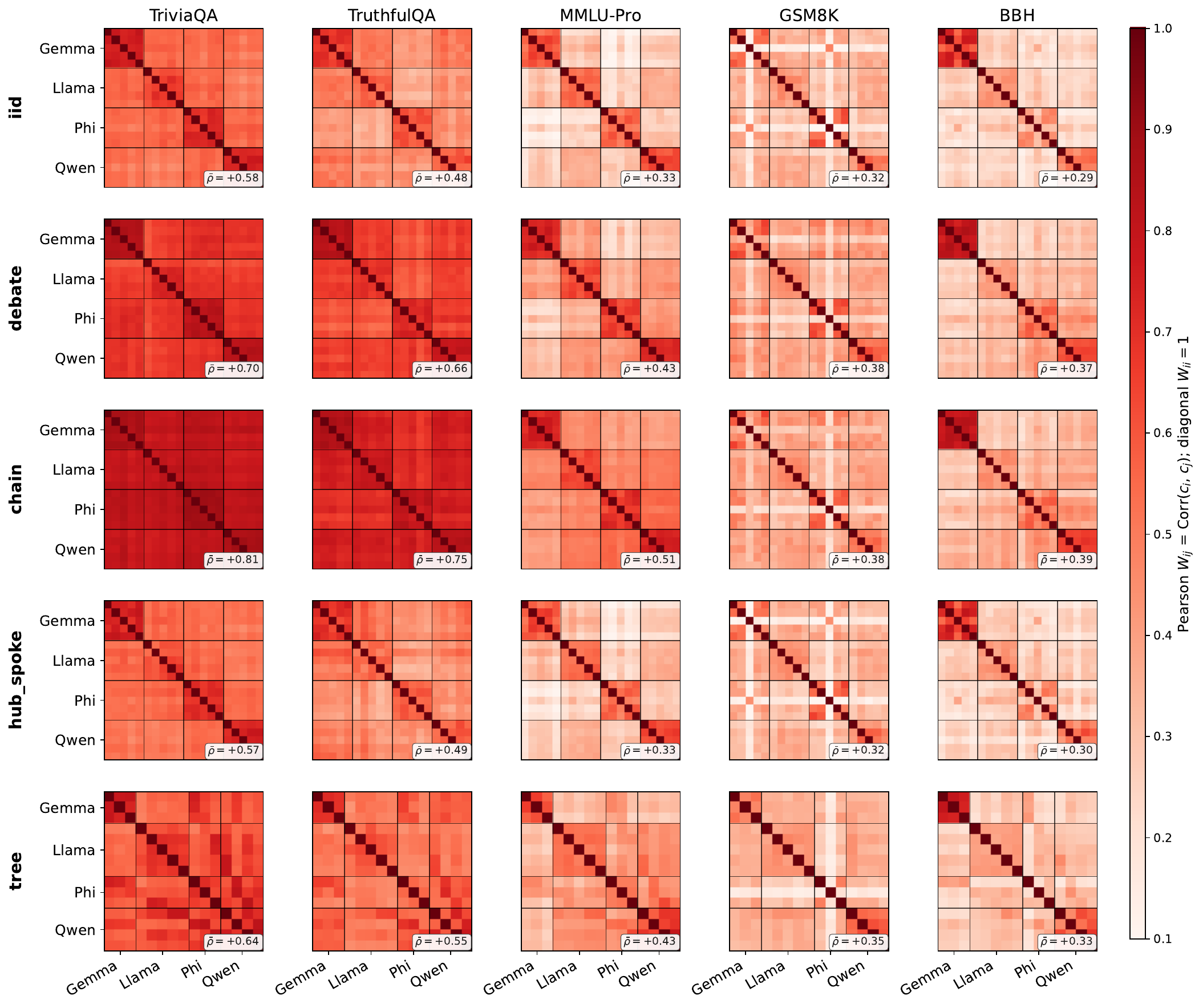}
  \caption{\textbf{Per-(topology, benchmark) agent error
  correlation $\W$.} Each panel is an $N \times N$ Pearson
  correlation of the binary error indicator $c_i$ across panel
  agents, with per-query mean residualisation to control for task
  difficulty. Agents are sorted by (backbone, role) so visual
  patterns are comparable across topologies.}
  \label{fig:correlation-per-cell}
\end{figure*}
Figure~\ref{fig:correlation-per-cell} shows that agent correctness correlations vary across model families, communication topologies, and benchmarks. 
Darker entries indicate that two agents tend to succeed or fail together. 
Within-family blocks, such as agents sharing the same Gemma, Llama, Phi, or Qwen backbone, are often darker, suggesting shared model-side blind spots. 
Comparing rows shows that topology also changes the dependence pattern: chain and debate often make the matrices darker than iid, indicating stronger communication-induced coupling. 
Comparing columns shows that this effect is benchmark-dependent: correlations are much stronger on TriviaQA and TruthfulQA than on MMLU-Pro, GSM8K, or BBH. 
These patterns show that multi-agent reliability depends jointly on heterogeneous LLM composition, communication structure, and task domain, motivating topology-aware calibration.

\section{Method Details}
\label{app:method_details}

\subsection{Node Feature Construction}
\label{con:details_of_node_feature}
Each node represents one agent. 
For both $G_x^T$ and $G_x^0$, we use the same 23-dimensional node feature schema:
\begin{equation*}
v_i(x)
=
[
s_i,\,
p_i,\,
r_i,\,
\mu_i,\,
\sigma_i^2,\,
m_i,\,
n^{-1},\,
a_i
]
\end{equation*}
The confidence score $s_i$ is computed from the agent's mean answer log probability. 
We clip the mean log probability to $[-10,0]$ and linearly map it to a normalized score. 
The plurality indicator $p_i$ is one when the agent's normalized answer matches the panel plurality answer. 
The vote rank $r_i$ is normalized by the number of distinct answer clusters in the panel. 
The correlation summaries $(\mu_i,\sigma_i^2,m_i)$ are the row mean, row variance, and row maximum of $|W_{ij}(x)|$ over other agents. 
For the observed tower, these summaries are computed from the post communication correlation matrix. 
For the iid counterfactual tower, they are computed from the iid correlation matrix. 
Finally, the answer embedding $a_i$ is obtained by encoding the agent's answer with Sentence-BERT~\citep{reimers2019sentencebertsentenceembeddingsusing} and reducing the embedding to 16 dimensions by PCA. 
Thus, each node records both the agent's individual contribution to the vote and its local dependence pattern with the rest of the panel.

\subsection{Instance-Conditional Correlation Estimation}
\label{con:corrlation_extimtion}
We estimate the correlation from the training split only. 
For each query, we retrieve the $k=20$ nearest training queries in Sentence-BERT embedding space and compute the empirical Pearson correlation of agent correctness over this local neighborhood. 
Pairs with nearly zero variance are assigned correlation zero. 
Validation and test labels are never used to construct these matrices.

In implementation, we keep an edge if either the agents are directly connected by communication or the absolute local correlation exceeds $0.05$. 
For the iid counterfactual graph, only the correlation criterion is used. 
This keeps the graph focused on meaningful dependence while preserving direct communication links. 
If no edge remains after thresholding, we use a fully connected graph without self loops to avoid degenerate isolated graphs.

\section{Implementation Details}
\label{app:experimental_setup}
\subsection{Benchmarks}
We evaluate on five English-language benchmarks chosen to span
distinct skill profiles: short-form factual recall (TriviaQA),
truthfulness in the face of common misconceptions (TruthfulQA), broad expert-level knowledge (MMLU-Pro), grade-school numerical reasoning (GSM8K), and a diverse battery of hard reasoning tasks (BBH).

\paragraph{TriviaQA~\citep{joshi2017triviaqalargescaledistantly}.}
A large-scale closed-book QA dataset of factoid trivia questions sourced from quiz league competitions. Gold answers are short strings, and an official multi-answer alias set is provided to accommodate surface-form variation at grading time.

\paragraph{TruthfulQA~\citep{lin2022truthfulqameasuringmodelsmimic}.}
$817$ questions designed to probe whether models reproduce common
human misconceptions or imitative falsehoods. We use the
open-ended generation setting, in which each question ships with
reference sets of correct and incorrect answers used to grade free
responses.

\paragraph{MMLU-Pro~\citep{wang2024mmluprorobustchallengingmultitask}.}
An enhanced version of MMLU with up to ten answer options per
question and stronger distractors than the original, covering
$14$ subject categories. Questions are presented in standard
A--J multiple-choice format and graded by the selected letter.

\paragraph{GSM8K~\citep{cobbe2021trainingverifierssolvemath}.}
Grade-school math word problems requiring multi-step arithmetic reasoning, each with a single numerical gold answer. Grading compares numerical equality of the model's parsed answer against the canonical gold value.

\paragraph{BBH~\citep{suzgun2022challengingbigbenchtaskschainofthought}.}
BIG-Bench Hard, a collection of $23$ subtasks drawn from BIG-Bench
on which prior LLMs underperformed humans. Subtasks vary in answer
format (multiple choice, yes/no, free-form), and grading dispatches
to a subtask-specific judge.

\subsection{Agent designs}
\paragraph{Backbones.}
We use four open-weight LLM backbones drawn from four distinct
model families: \textsc{Qwen3-8B}~\citep{qwen2025qwen25}, \textsc{Llama-3.1-8B}~\citep{meta2024llama3}, \textsc{Gemma-3-12B}~\citep{gemma_2025}, and \textsc{Phi-4}~\citep{phi_2024}. Using four independently pretrained families is intended to expose realistic disagreement: agents sometimes fail in correlated ways because they share training data or recipe, and sometimes disagree because they come from different pipelines.

\paragraph{Prompting roles.}
Each backbone is paired with one of five atomic single-pass
reasoning roles, each adapted verbatim from its source paper and summarized in Appendix~\ref{app:prompt}:
\textsc{direct} (zero-shot, no reasoning shown),
\textsc{cot}~\citep{wei2023chainofthoughtpromptingelicitsreasoning},
\textsc{plan\_solve}~\citep{wang2023planandsolvepromptingimprovingzeroshot},
\textsc{step\_back}~\citep{zheng2024stepbackevokingreasoning}, and
\textsc{analogical}~\citep{yasunaga2024largelanguagemodelsanalogical}. Restricting each agent to a single LLM call and a single role keeps the role axis and the topology axis orthogonal, so any multi-round or multi-agent behavior comes from the topology rather than from a role internally hiding a sub-population.

\paragraph{Canonical population.}
An \emph{agent} is a (backbone, prompting role) pair. The full
$4 \times 5$ cross product yields $20$ deterministically ordered atomic agents that constitute our canonical population, and the \emph{same} population is reused across every topology so that any cross-topology difference can be attributed to communication structure rather than to a different mix of agents.

\subsection{Communication Topology Design}
\label{app:topology-design}

We compare five communication topologies that span the spectrum
from no communication to fully connected, and from peer-to-peer to
hierarchical aggregation. All topologies share the same agent
population, the same per-agent sampling temperature,
the same maximum decode length, and the same
plurality-vote aggregator at the end. Each topology is run with $3$ rollouts per query for reproducibility.

\paragraph{\textsc{iid} (no communication).}
The $N$ agents answer the query independently and in parallel.
This is the standard self-consistency setting and serves as the counterfactual baseline against which the four communicating topologies are compared. The iid rollout is also the source of the counterfactual graph $G^{0}_{x}$ used by \methodname{}.

\paragraph{\textsc{debate} (two-round full mesh).}
Round~1 is identical to iid. In round~2, every agent receives a formatted block listing all other agents' round-1 answers and is invited to revise. The final panel is the set of round-2 answers, so every agent has been exposed to every peer exactly once before committing.

\paragraph{\textsc{chain} (sequential pipeline).}
The $N$ agents are arranged in a per-rollout shuffled order
$a_{1} \to a_{2} \to \cdots \to a_{N}$. Agent $a_{i}$ sees the
formatted answers of $a_{1}, \ldots, a_{i-1}$ before producing its own. The final panel is the set of all $N$ answers; the last agent has the largest peer context, the first agent has none.

\paragraph{\textsc{hub-spoke} (centralized aggregation).}
The $N{-}1$ spoke agents answer the query independently and in
parallel. The hub then sees a formatted block of all spoke answers labelled as ``worker'' contributions and produces its own answer. The final panel includes the hub and all spokes, with the hub's answer typically dominating the plurality vote.

\paragraph{\textsc{tree} (hierarchical aggregation).}
A complete binary tree of depth $L$ aggregates leaf answers upward. At the bottom layer, $2^{L}$ leaf agents answer independently. Each internal node sees the answers of its two children and produces its own answer; the root produces the panel's most informed answer. All nodes' answers are retained for the panel. With our canonical $N{=}15$ subset, the tree has depth $L{=}3$ and shape $8{+}4{+}2{+}1$.

\subsection{Baselines}
We compare \methodname{} against four families of baselines. The first family applies classical post-hoc calibration to the raw plurality vote share. The second asks an LLM judge to score the panel directly. The third trains a learned calibration head on validation data. The fourth reports ranking-only uncertainty scores that do not target
calibrated probabilities.

\subsubsection{Post-hoc plurality calibrators}
\label{app:baselines-posthoc}

Methods in this family use the plurality vote share as their raw
confidence signal~\citep{kuncheva2004combining}. This is the fraction
of agents that support the panel's most-voted answer. The four methods
differ only in how this scalar is reshaped into a final probability.
\emph{Plurality share} uses the raw value with no learnable parameters
and serves as the simplest calibration reference.
\emph{$+$ Platt scaling}~\citep{platt1999probabilistic} fits a
parametric logistic mapping on the validation split.
\emph{$+$ Isotonic regression}~\citep{zadrozny2002transforming}
instead fits a non-parametric monotone mapping. It can correct
non-sigmoidal miscalibration but introduces more variance.
\emph{$+$ Scaling-binning}~\citep{kumar2019verified} chains a
parametric scaler with empirical bin-mean replacement and provides
finite-sample ECE guarantees.

\subsubsection{LLM-elicited confidence estimators}
\label{app:baselines-llm}

These baselines test whether panel correctness can be inferred from the final agent responses alone. We prompt an LLM judge to return a probability that the plurality answer is correct. The judge sees the question, the agent answers, and the plurality answer, but no graph or per-token statistics. Full prompts appear in Appendix~\ref{app:prompt}. \emph{LLM-Cal (no topology)} queries the judge directly. \emph{LLM-Cal ($+$topology)} additionally injects a one-line description of the panel's communication topology into the prompt. This isolates whether the judge can exploit topology
information when given it explicitly. \emph{Collaborative
Calibration}~\citep{yang2024confidence} prompts the judge to silently simulate a multi-expert deliberation before emitting a single consensus probability.

\subsubsection{Trained calibrators}
\label{app:baselines-trained}

These baselines train a calibration head on validation panels and target plurality-answer correctness. To make the comparison fair, all three trained baselines receive the same per-benchmark post-hoc protocol as our method. \emph{Scalar~$+$~GBT}~\citep{ke2017lightgbm}
is a strong feature-based baseline. It trains a gradient-boosted decision tree on hand-crafted scalar panel statistics, with no relational encoding. \emph{GraphCal}~\citep{li2025graph} adapts a
graph-based calibrator to the panel setting. It encodes the observed agent graph with a GCN~\citep{kipf2017semisupervisedclassificationgraphconvolutional}, but it omits the counterfactual iid view, the hyperedge stream, and the failure-correlation edges used by \methodname{}.
\emph{DiscoUQ-LLM}~\citep{jiang2026discouq} is a disagreement-feature baseline. It trains a head on a small set of panel-level disagreement summaries.
\begin{table*}[!t]
\centering
\setlength{\tabcolsep}{3pt}
\resizebox{\textwidth}{!}{%
\begin{tabular}{l c c c c c c c c c c c c}
\toprule
& \multicolumn{2}{c}{\textbf{TriviaQA}}
& \multicolumn{2}{c}{\textbf{TruthfulQA}}
& \multicolumn{2}{c}{\textbf{MMLU-Pro}}
& \multicolumn{2}{c}{\textbf{GSM8K}}
& \multicolumn{2}{c}{\textbf{BBH}}
& \multicolumn{2}{c}{\textbf{Mean}} \\
\cmidrule(lr){2-3}\cmidrule(lr){4-5}\cmidrule(lr){6-7}\cmidrule(lr){8-9}\cmidrule(lr){10-11}\cmidrule(lr){12-13}
\textbf{Method}
  & AUROC$\uparrow$ & AUARC$\uparrow$
  & AUROC$\uparrow$ & AUARC$\uparrow$
  & AUROC$\uparrow$ & AUARC$\uparrow$
  & AUROC$\uparrow$ & AUARC$\uparrow$
  & AUROC$\uparrow$ & AUARC$\uparrow$
  & AUROC$\uparrow$ & AUARC$\uparrow$ \\
\midrule
Plurality vote     & 81.89 {\scriptsize $\pm$ 0.79} & \textbf{93.09 {\scriptsize $\pm$ 1.14}} & 71.16 {\scriptsize $\pm$ 3.89} & 31.30 {\scriptsize $\pm$ 1.22} & 61.78 {\scriptsize $\pm$ 0.80} & 53.02 {\scriptsize $\pm$ 0.40} & 84.01 {\scriptsize $\pm$ 2.22} & \textbf{97.87 {\scriptsize $\pm$ 0.22}} & 66.15 {\scriptsize $\pm$ 1.49} & 84.22 {\scriptsize $\pm$ 0.46} & 72.99 {\scriptsize $\pm$ 1.52} & 71.90 {\scriptsize $\pm$ 0.07} \\
Answer entropy     & 80.19 {\scriptsize $\pm$ 0.67} & 92.82 {\scriptsize $\pm$ 1.11} & 70.83 {\scriptsize $\pm$ 2.97} & 31.36 {\scriptsize $\pm$ 0.93} & 57.92 {\scriptsize $\pm$ 0.98} & 51.01 {\scriptsize $\pm$ 0.70} & 82.60 {\scriptsize $\pm$ 0.99} & 97.72 {\scriptsize $\pm$ 0.08} & 67.98 {\scriptsize $\pm$ 1.30} & 84.94 {\scriptsize $\pm$ 0.52} & 71.90 {\scriptsize $\pm$ 1.11} & 71.57 {\scriptsize $\pm$ 0.06} \\
Avg-logprob        & 69.83 {\scriptsize $\pm$ 2.81} & 86.96 {\scriptsize $\pm$ 0.74} & 60.31 {\scriptsize $\pm$ 0.41} & 27.65 {\scriptsize $\pm$ 1.01} & 43.65 {\scriptsize $\pm$ 2.08} & 27.33 {\scriptsize $\pm$ 0.40} & 57.87 {\scriptsize $\pm$ 2.32} & 75.41 {\scriptsize $\pm$ 0.48} & 46.11 {\scriptsize $\pm$ 1.90} & 49.77 {\scriptsize $\pm$ 1.47} & 55.56 {\scriptsize $\pm$ 0.91} & 53.42 {\scriptsize $\pm$ 0.82} \\
DAE                & 80.19 {\scriptsize $\pm$ 0.66} & 92.82 {\scriptsize $\pm$ 1.11} & 70.85 {\scriptsize $\pm$ 3.01} & 31.34 {\scriptsize $\pm$ 0.95} & 57.91 {\scriptsize $\pm$ 0.94} & 51.01 {\scriptsize $\pm$ 0.72} & 82.61 {\scriptsize $\pm$ 1.00} & 97.72 {\scriptsize $\pm$ 0.08} & 67.96 {\scriptsize $\pm$ 1.31} & 84.94 {\scriptsize $\pm$ 0.53} & 71.91 {\scriptsize $\pm$ 1.12} & 71.57 {\scriptsize $\pm$ 0.05} \\
MATU               & 71.23 {\scriptsize $\pm$ 2.55} & 89.73 {\scriptsize $\pm$ 0.33} & 59.80 {\scriptsize $\pm$ 2.57} & 26.88 {\scriptsize $\pm$ 1.82} & 61.39 {\scriptsize $\pm$ 0.96} & 52.20 {\scriptsize $\pm$ 0.24} & 56.68 {\scriptsize $\pm$ 1.32} & 95.45 {\scriptsize $\pm$ 0.19} & 48.81 {\scriptsize $\pm$ 2.91} & 73.89 {\scriptsize $\pm$ 0.19} & 59.58 {\scriptsize $\pm$ 1.68} & 67.63 {\scriptsize $\pm$ 0.42} \\
\midrule
\rowcolor[RGB]{222,230,241}
\textbf{\methodname{} (ours)}
                   & \textbf{86.12 {\scriptsize $\pm$ 0.78}} & 93.02 {\scriptsize $\pm$ 0.26}
                   & \textbf{79.66 {\scriptsize $\pm$ 2.43}} & \textbf{38.76 {\scriptsize $\pm$ 1.44}}
                   & \textbf{77.74 {\scriptsize $\pm$ 0.75}} & \textbf{61.54 {\scriptsize $\pm$ 0.88}}
                   & \textbf{84.04 {\scriptsize $\pm$ 3.55}} & 97.51 {\scriptsize $\pm$ 0.51}
                   & \textbf{90.48 {\scriptsize $\pm$ 0.81}} & \textbf{91.52 {\scriptsize $\pm$ 0.74}}
                   & \textbf{83.61 {\scriptsize $\pm$ 1.34}} & \textbf{76.47 {\scriptsize $\pm$ 0.37}} \\
\bottomrule
\end{tabular}%
}
\caption{\textbf{Heuristic UQ baselines and \methodname{}: per-benchmark
AUROC and AUARC.} Mean $\pm$ std over 3 rollouts, averaged
across the 5 topologies. ECE not applicable to the five
ranking-only methods; \methodname{}'s ECE is in
Table~\ref{tab:main}. \methodname{} AUROC matches
Table~\ref{tab:main}.}
\label{tab:heuristic-uq}
\end{table*}

\subsubsection{Ranking-only UQ scores}
\label{app:baselines-ranking}

This family does not produce calibrated probabilities. It only
yields scalar uncertainty estimates that can rank panels by
reliability, so we report only AUROC and AUARC for these methods. \emph{Answer entropy}~\citep{kuhn2023semanticuncertaintylinguisticinvariances} is the Shannon entropy of the panel's distribution over distinct agent answers. \emph{Average log probability}~%
\citep{kadavath2022languagemodelsmostlyknow} averages each agent's mean per-token log-probability across the panel. It uses only signals that the decoder already provides. \emph{DiverseAgentEntropy}~%
\citep{feng2025rethinkingllmuncertaintymultiagent} extends answer entropy by first diversifying the agents that produce the answer distribution. \emph{MATU}~\citep{chen2026responsecountsquantifyinguncertainty} arranges the panel's per-agent answer distribution into a low-rank tensor and uses the \textsc{PARAFAC2} reconstruction residual as the uncertainty score.

\subsection{Training Details}
\paragraph{Data splits.}
For each of the five benchmarks we sample $500$ problems with a fixed seed and deterministically split them $60{:}20{:}20$ into train, validation, and test partitions. The same split is reused across every topology, every method, and every seed, so a question is never seen in training under one topology and held out under another. Because each problem is rolled out through every topology with three independent rollouts, the resulting panel pool
contains roughly $37{,}500$ panels in total. The training split fits the encoder and the calibration head, the validation split fits the post-hoc calibrators and selects model checkpoints, and the test split is used only for final reporting. For the leave-one-topology-out experiments the same per-benchmark splits are kept fixed and only the topology used during training is varied.

\paragraph{Optimization.}
We train the encoder for $15$ epochs with AdamW, an initial
learning rate of $2\!\times\!10^{-3}$, weight decay of
$3\!\times\!10^{-4}$, and a cosine learning-rate schedule. Each batch contains $128$ panels and gradients are clipped to a global norm of $1.0$. The training loss combines binary cross-entropy on plurality-answer correctness with a Brier auxiliary term weighted by $0.4$, and the target is label-smoothed with $\alpha=0.05$ to prevent the model from collapsing to extreme probabilities on high-agreement panels. Model selection on the validation split uses the composite score $\text{AUROC}-0.5\cdot\text{ECE}$, with an early-stopping patience of $5$ epochs. We train an ensemble of $10$
random seeds and average the predicted probabilities, then apply a per-benchmark Beta calibration and scaling-binning post-hoc and blend the two outputs with equal weight.

\paragraph{Computational cost.}
The bulk of the wall-clock cost in our pipeline lies in generating the multi-agent panels with vLLM-served open-weight LLMs, which we cache and reuse across all methods. Training \methodname{} on top of the cached panels is comparatively cheap: a single $15$-epoch training run completes in roughly $5$--$10$ minutes on a single
NVIDIA A100, and the full $10$-seed ensemble plus per-benchmark post-hoc fits in well under two hours on the same hardware.

\section{Additional Calibration and Uncertainty Results}
\label{app:additional_results}

\subsection{Ranking-Only UQ Comparison}
\label{con:Ranking-Only UQ Comparison}

Table~\ref{tab:heuristic-uq} compares \methodname{} with heuristic uncertainty baselines across benchmarks. 
Plurality vote and answer entropy rely only on the final answer distribution, while log-probability and other uncertainty scores do not explicitly model agent dependencies. 
\methodname{} achieves the best mean AUROC and AUARC, showing that topology-aware dependency features provide a stronger signal for estimating whether the panel answer is correct.

\subsection{Brier Score Comparison}
\label{con:brier_score}
\begin{table*}[!t]
\centering
\setlength{\tabcolsep}{5pt}
\resizebox{0.7\textwidth}{!}{%
\begin{tabular}{l c c c c c c}
\toprule
\textbf{Method}
  & \textbf{TriviaQA} & \textbf{TruthfulQA} & \textbf{MMLU-Pro}
  & \textbf{GSM8K} & \textbf{BBH} & \textbf{Mean} \\
\midrule
\rowcolor{gray!20}
\multicolumn{7}{c}{\textit{Post-hoc plurality calibrators}}\\
Plurality share     & 12.12 {\scriptsize $\pm$ 0.23} & 24.52 {\scriptsize $\pm$ 0.90} & 25.01 {\scriptsize $\pm$ 0.18} &  9.61 {\scriptsize $\pm$ 0.21} & 23.89 {\scriptsize $\pm$ 0.11} & 19.03 {\scriptsize $\pm$ 0.19} \\
\;+\;Platt          & 10.91 {\scriptsize $\pm$ 0.06} & 16.36 {\scriptsize $\pm$ 0.76} & 24.30 {\scriptsize $\pm$ 0.19} &  4.37 {\scriptsize $\pm$ 0.18} & 17.97 {\scriptsize $\pm$ 0.02} & 14.78 {\scriptsize $\pm$ 0.21} \\
\;+\;Isotonic       & 10.89 {\scriptsize $\pm$ 0.18} & 16.45 {\scriptsize $\pm$ 0.93} & 24.76 {\scriptsize $\pm$ 0.42} &  4.50 {\scriptsize $\pm$ 0.27} & 17.73 {\scriptsize $\pm$ 0.29} & 14.87 {\scriptsize $\pm$ 0.35} \\
\;+\;Scaling-bin.   & 12.29 {\scriptsize $\pm$ 0.09} & 16.06 {\scriptsize $\pm$ 0.30} & 24.21 {\scriptsize $\pm$ 0.02} &  4.54 {\scriptsize $\pm$ 0.01} & 17.98 {\scriptsize $\pm$ 0.11} & 15.02 {\scriptsize $\pm$ 0.03} \\
\midrule
\rowcolor{gray!20}
\multicolumn{7}{c}{\textit{LLM-elicited confidence estimators}}\\
LLM-Cal (no topo)   & 10.11 {\scriptsize $\pm$ 0.09} & 46.90 {\scriptsize $\pm$ 0.34} & 35.46 {\scriptsize $\pm$ 1.36} & 10.19 {\scriptsize $\pm$ 0.04} & 19.37 {\scriptsize $\pm$ 0.69} & 24.41 {\scriptsize $\pm$ 0.45} \\
LLM-Cal (+topo)     & 10.75 {\scriptsize $\pm$ 0.07} & 46.51 {\scriptsize $\pm$ 0.24} & 34.60 {\scriptsize $\pm$ 0.52} &  8.96 {\scriptsize $\pm$ 0.56} & 19.40 {\scriptsize $\pm$ 0.09} & 24.04 {\scriptsize $\pm$ 0.04} \\
Collab.~Cal.        & 10.32 {\scriptsize $\pm$ 0.32} & 44.21 {\scriptsize $\pm$ 0.63} & 33.71 {\scriptsize $\pm$ 0.97} &  8.41 {\scriptsize $\pm$ 0.74} & 18.26 {\scriptsize $\pm$ 0.33} & 22.98 {\scriptsize $\pm$ 0.17} \\
\midrule
\rowcolor{gray!20}
\multicolumn{7}{c}{\textit{Trained calibrators}}\\
Scalar~+~GBT        & 11.43 {\scriptsize $\pm$ 0.62} & 17.49 {\scriptsize $\pm$ 1.05} & 25.96 {\scriptsize $\pm$ 0.75} &  4.87 {\scriptsize $\pm$ 0.05} & 18.11 {\scriptsize $\pm$ 0.05} & 15.57 {\scriptsize $\pm$ 0.22} \\
GraphCal            & 12.29 {\scriptsize $\pm$ 0.35} & 28.32 {\scriptsize $\pm$ 0.77} & 26.28 {\scriptsize $\pm$ 0.08} &  9.25 {\scriptsize $\pm$ 0.03} & 20.18 {\scriptsize $\pm$ 0.02} & 19.26 {\scriptsize $\pm$ 0.10} \\
DiscoUQ-LLM         & 10.47 {\scriptsize $\pm$ 0.19} & 15.31 {\scriptsize $\pm$ 0.61} & 24.02 {\scriptsize $\pm$ 0.03} &  4.46 {\scriptsize $\pm$ 0.09} & 17.57 {\scriptsize $\pm$ 0.14} & 14.37 {\scriptsize $\pm$ 0.16} \\
\midrule
\rowcolor[RGB]{222,230,241}
\textbf{\methodname{} (ours)}
                    & \textbf{ 9.55 {\scriptsize $\pm$ 0.80}} & \textbf{13.57 {\scriptsize $\pm$ 0.53}} & \textbf{19.26 {\scriptsize $\pm$ 0.46}} & \textbf{ 4.33 {\scriptsize $\pm$ 0.22}} & \textbf{ 9.26 {\scriptsize $\pm$ 0.48}} & \textbf{11.19 {\scriptsize $\pm$ 0.50}} \\
% \midrule
% \multicolumn{7}{l}{\textit{Upper bound (uses test labels)}}\\
% Oracle isotonic     &  9.38 {\scriptsize $\pm$ 0.49} & 14.47 {\scriptsize $\pm$ 0.55} & 22.65 {\scriptsize $\pm$ 0.31} &  3.78 {\scriptsize $\pm$ 0.10} & 16.05 {\scriptsize $\pm$ 0.00} & 13.26 {\scriptsize $\pm$ 0.09} \\
\bottomrule
\end{tabular}%
}
\caption{\textbf{Brier score ($\downarrow$) per benchmark.}
Mean $\pm$ std over 3 rollouts, averaged across the 5
topologies. Reported only for methods that produce a probability
natively. \methodname{} attains the lowest Brier on every
benchmark and the lowest Mean.}
\label{tab:brier}
\end{table*}

Table~\ref{tab:brier} compares probability calibration using Brier score, where lower values indicate better calibrated confidence. 
The baselines cover three categories: post-hoc plurality calibrators, LLM-elicited confidence estimators, and trained calibrators. 
\methodname{} achieves the lowest Brier score on every benchmark and the best mean score overall, reducing the average Brier score from the strongest baseline, DiscoUQ-LLM, from 14.37 to 11.19. 
This suggests that explicitly modeling agent dependencies and communication topology provides a stronger calibration signal than calibrating vote share, eliciting verbalized confidence, or using hand-crafted disagreement features alone.
\begin{table*}[!t]
\centering
\setlength{\tabcolsep}{4pt}
\resizebox{0.65\textwidth}{!}{%
\begin{tabular}{l c c c c c c}
\toprule
\textbf{Strategy} & \textbf{TriviaQA} & \textbf{TruthfulQA}
                  & \textbf{MMLU-Pro} & \textbf{GSM8K}
                  & \textbf{BBH} & \textbf{Mean} \\
\midrule
\rowcolor{gray!20}
\multicolumn{7}{c}{\textit{Fixed topology}}\\
iid                                      & 83.67 & 20.00 & 41.50 & 94.67 & 72.51 & 62.41 \\
debate                                   & 84.01 & 23.67 & 46.94 & 94.67 & 75.95 & 64.98 \\
chain                                    & 82.65 & 20.67 & 46.26 & 94.00 & 75.95 & 63.83 \\
hub-spoke                                & 84.01 & 20.33 & 42.86 & 95.33 & 72.85 & 63.02 \\
tree                                     & 82.99 & 19.67 & 41.50 & 95.00 & 70.10 & 61.80 \\
Per-bench best fixed         & 84.01 & 23.67 & 46.94 & 95.33 & 75.95 & 65.18 \\
\midrule
\rowcolor{gray!20}
\multicolumn{7}{c}{\textit{Selection w/o learned confidence}}\\
Majority over topologies                 & 84.69 & 19.00 & 40.82 & 95.33 & 75.26 & 62.95 \\
Highest plurality share                  & 84.35 & 22.33 & 44.56 & 95.33 & 78.69 & 64.98 \\
Highest mean log-prob                    & 84.69 & 22.33 & 40.82 & 94.67 & 74.23 & 63.29 \\
\midrule
\rowcolor{gray!20}
\multicolumn{7}{c}{\textit{Confidence-routed}}\\
\rowcolor[RGB]{222,230,241}
\textbf{\textsc{CAGE-Select} (ours)}                     & \textbf{84.69}
                                          & \textbf{24.00}
                                          & \textbf{50.34}
                                          & \textbf{95.33}
                                          & \textbf{81.79}
                                          & \textbf{67.23} \\
\midrule
Oracle topology            & 88.10 & 31.67 & 65.31 & 96.33 & 86.25 & 73.43 \\
\bottomrule
\end{tabular}%
}
\caption{\textbf{Confidence-routed topology selection.} Mean
accuracy on the $1{,}479$ matched-$N$ test groups (one per query,
five candidate panels). \emph{Per-bench best fixed} picks each
benchmark's best topology on validation.}
\label{tab:cage-select}
\end{table*}

\section{Selection, Generalization, and Robustness}
\label{app:robustness}

\subsection{\select{}: Confidence-Routed Topology Selection}
\label{con:performance_cage_select}

Table~\ref{tab:cage-select} evaluates whether the calibrated
confidence produced by \methodname{} can be used to select, for each query, the communication topology whose panel answer is most likely to be correct. We refer to this routing procedure as \select{}.

\paragraph{Why per-query routing.}
The main results show that no single communication topology is best on every query. Some questions are answered most reliably under iid, where independent agents avoid the herding that arises when they communicate. Others benefit from debate or hub-spoke, where peer exchange resolves ambiguity that a lone agent would miss. A \emph{fixed}-topology system is therefore always sub-optimal in
expectation: it commits, at design time, to a single communication structure regardless of which structure is appropriate for the specific input. \select{} replaces that design-time commitment with a run-time choice driven by panel-level confidence.

\paragraph{Routing procedure.}
For each query $x$, \select{} runs the panel under all candidate topologies, producing a plurality answer $\hat{y}(x,T)$ and a calibrated correctness probability $\hat{p}(x,T)$ from \methodname{} for each $T$. It then returns the answer of $T^{\star}(x) = \arg\max_{T} \hat{p}(x,T)$. Because $\hat{p}$ is trained as a panel-level correctness probability with the same target across topologies, the values are directly comparable and the routing reduces to a single $\arg\max$ over a small candidate pool. We compare \select{} against each fixed topology, an oracle that always picks the topology carrying the correct answer, and heuristic routers that select by plurality share or mean per-token log-probability. \select{} consistently beats every fixed topology
and every heuristic rule and closes a substantial fraction of the oracle gap. The heuristics fail in characteristic ways: plurality share is fooled by communication-induced herding that inflates agreement without improving accuracy, and mean log-probability rewards confident generation regardless of whether the panel agrees. A calibrated panel-level confidence is needed to navigate both
failure modes, and the same head that flags unreliable panels can therefore double as an inference-time selector that turns topology choice into a per-query decision.

% Instead of using a fixed topology for all queries, \select{} runs candidate panels under multiple topologies and selects the answer with the highest \methodname{} confidence. 
%  We found it   outperform fixed-topology baselines and heuristic selection rules such as highest plurality share or highest mean log-probability. 
% The improvement shows that calibrated confidence is useful not only for reliability estimation, but also for choosing which topology-induced answer to trust on each query.

\subsection{Panel-Size Robustness}
\label{con:generalization_to_com_top}
\begin{table}[!t]
\centering
\setlength{\tabcolsep}{4pt}
\resizebox{\columnwidth}{!}{%
\begin{tabular}{l c c c c c c}
\toprule
& \multicolumn{2}{c}{\textbf{GraphCal}}
& \multicolumn{2}{c}{\textbf{DiscoUQ-LLM}}
& \multicolumn{2}{c}{\textbf{\methodname{} (ours)}}\\
\cmidrule(lr){2-3}\cmidrule(lr){4-5}\cmidrule(lr){6-7}
\textbf{Held-out topology} & AUROC & AUARC & AUROC & AUARC & AUROC & AUARC\\
\midrule
iid          & $67.67$ & $69.63$ & $74.89$ & $71.45$ & \textbf{$84.15$} & \textbf{$75.23$} \\
debate       & $69.95$ & $71.59$ & $70.32$ & $71.67$ & \textbf{$80.49$} & \textbf{$76.60$} \\
chain        & $67.98$ & $70.11$ & $66.97$ & $70.11$ & \textbf{$82.55$} & \textbf{$77.32$} \\
hub-spoke    & $70.80$ & $71.87$ & $73.51$ & $72.44$ & \textbf{$80.12$} & \textbf{$74.71$} \\
tree         & $73.58$ & $70.63$ & $75.66$ & $70.72$ & \textbf{$83.74$} & \textbf{$74.92$} \\
\midrule
Mean         & $70.00$ & $70.77$ & $72.27$ & $71.28$ & \textbf{82.21} & \textbf{75.76} \\
\bottomrule
\end{tabular}%
}
\caption{\textbf{LOTO per-held-out breakdown} (percent, AUROC
and AUARC). Each row trains on the other four topologies and
tests on the held-out fifth. The Mean row reproduces Table~\ref{tab:main-loto}'s Mean
column.}
\label{tab:loto}
\end{table}

Table~\ref{tab:loto} evaluates leave-one-topology-out generalization, where each row trains the calibrator on four communication topologies and tests it on the held-out topology. 
\methodname{} consistently outperforms GraphCal and DiscoUQ-LLM in both AUROC and AUARC across all held-out topologies, showing that its graph-based dependency modeling generalizes beyond the topologies observed during training. 
This suggests that \methodname{} learns reusable structural signals of agent dependence rather than merely fitting topology-specific patterns.

\subsection{Main Structural Findings Are Robust to Panel Size}
\label{con:pannel_size}

\textbf{\begin{table}[!t]
\centering
\setlength{\tabcolsep}{6pt}
\resizebox{\columnwidth}{!}{%
\begin{tabular}{l c c c}
\toprule
\textbf{Quantity} & $\bm{N{=}10}$ & $\bm{N{=}20}$ & $\bm{\Delta}$ \\
\midrule
\multicolumn{4}{l}{\textit{Accuracy anchors (\%)}}\\
Single-agent baseline           & 54.13  & 54.72  & +0.59 \\
iid plurality                   & 60.25  & 64.34  & +4.09 \\
Best topology per cell          & 63.00  & 66.40  & +3.41 \\
Any-agent-correct oracle        & 84.68  & 88.08  & +3.39 \\
Ensembling gain (iid $-$ single) & +6.12  & +9.62  & +3.50 \\
Communication gain (best $-$ iid)& +2.75  & +2.06  & -0.69 \\
Aggregation gap (oracle $-$ best)& +21.69 & +21.67 & -0.01 \\
\midrule
\multicolumn{4}{l}{\textit{OLS $\beta$ on $\W_{ij}$ (pp)}}\\
Same backbone family             & +10.17 & +10.36 & +0.19 \\
Same prompting role              & +1.02  & +0.89  & -0.14 \\
\texttt{is\_chain}               & +18.38 & +17.87 & -0.51 \\
\texttt{is\_debate}              & +12.19 & +11.76 & -0.43 \\
\texttt{is\_tree}                & +6.20  & +6.76  & +0.55 \\
\texttt{is\_hub\_spoke}          & +1.46  & +1.15  & -0.31 \\
\midrule
\multicolumn{4}{l}{\textit{Peak regression rate (\%)}}\\
chain / TruthfulQA               & 32.45  & 31.73  & -0.72 \\
tree / TruthfulQA                & 31.18  & 30.52  & -0.66 \\
\bottomrule
\end{tabular}%
}
\caption{\textbf{$N$-scaling robustness.} Headline quantities
under matched-$N{=}10$ vs.\ natural-$N{=}20$. Structural findings
move by less than $\pm 1$\,pp.}
\label{tab:n-scaling-side-by-side}
\end{table}
}

Table~\ref{tab:n-scaling-side-by-side} evaluates whether our structural findings are robust to the number of agents in the panel by comparing matched \(N=10\) panels with the natural \(N=20\) setting. 
Increasing the panel size improves standard accuracy anchors, such as iid plurality accuracy and the any-agent-correct oracle, but the key structural quantities remain stable. 
In particular, the OLS coefficients on pairwise agent dependence, including same backbone family and topology-induced effects such as chain and debate, change by less than about one percentage point. 
The peak regression rates on TruthfulQA are also nearly unchanged. 
These results suggest that the observed population-side dependence and topology-induced coupling are not artifacts of a particular panel size.

\section{Prompt Design}
\label{app:prompt}
\paragraph{Agent role prompts}
Each agent in a panel is parameterized by one of five atomic reasoning
roles whose prompts are taken from
the originating papers; a single shared format guard is appended so
that downstream parsing of the "Answer:~$\langle X\rangle$" terminator
is uniform across roles and benchmarks. A role is restricted to one
LLM call and may not internally simulate multiple personas, which
keeps the role axis orthogonal to the topology axis. Agent prompting roles used to generate panels are shown in Figure~\ref{fig:appendix:role-prompts}.
% Figure ref{fig:appendix:role-prompts}
\begin{figure*}[h]
\centering
\small

\begin{promptbox}{Agent prompting roles}
\textbf{Shared format guard} \textit{(appended to every role below):}\\[2pt]
\texttt{Output the final answer at the end as exactly one line:
"Answer: <your short answer>"}

\vspace{6pt}\hrule\vspace{4pt}

\textbf{Role 1: \textsc{direct}} (zero-shot baseline)\\[2pt]
\texttt{Answer the question directly. Do not show reasoning.}

\vspace{6pt}

\textbf{Role 2: \textsc{cot}}~\citep{wei2023chainofthoughtpromptingelicitsreasoning}\\[2pt]
\texttt{Let's think step by step. Reason through the problem,
then commit to a final answer.}

\vspace{6pt}

\textbf{Role 3: \textsc{plan\_solve}}~\citep{wang2023planandsolvepromptingimprovingzeroshot}\\[2pt]
\texttt{First, understand the problem and devise a brief plan in
2--4 steps. Then carry out the plan to solve the problem.}

\vspace{6pt}

\textbf{Role 4: \textsc{step\_back}}~\citep{zheng2024stepbackevokingreasoning}\\[2pt]
\texttt{Take a step back. State the high-level concept, principle,
or category that this problem falls under. Then use that principle
to solve the specific problem.}

\vspace{6pt}

\textbf{Role 5: \textsc{analogical}}~\citep{yasunaga2024largelanguagemodelsanalogical}\\[2pt]
\texttt{Recall 2--3 analogous problems you have seen before.
Briefly describe each in one sentence. Then use what you learned
from those analogies to solve this problem.}
\end{promptbox}

\caption{\textbf{Agent prompting roles used to generate panels.}
Five atomic reasoning roles parameterize each agent, adapted
verbatim from the cited source papers. A role must not internally simulate multiple
personas or multiple LLM calls; this keeps role and topology
orthogonal.}
\label{fig:appendix:role-prompts}
\end{figure*}

\paragraph{LLM-elicited baseline prompts}
For completeness we report the prompt used by the LLM-Cal baseline
(Section~\ref{sec:baselines}), the zero-shot LLM-elicited calibrator that
asks a frozen LLM to map (question, panel answers, plurality) to a
correctness probability. The optional \texttt{+topo} variant
additionally injects a one-line description of the panel's
communication topology; the exact label mapping is shown in the same
box.  The prompt used for LLM-Cal is shown in Figure~\ref{fig:appendix:llmcal-prompt}.
% Figure ref{fig:appendix:llmcal-prompt}
\begin{figure*}[t]
\centering
\small

\begin{promptbox2}{LLM-Cal (zero-shot LLM-elicited calibrator)}
\textbf{System prompt:}\\[2pt]
\texttt{You are a calibration assistant. Given a question and N
candidate answers from N language model agents, plus the plurality
(most-voted) answer, estimate the probability that the plurality
answer is correct. Respond with a single decimal number between 0
and 1, no other text.}

\vspace{6pt}\hrule\vspace{4pt}

\textbf{User prompt template:}\\[2pt]
\texttt{Question: \{query\}\\
\\
Panel topology: \{topology\_description\}\;\textit{\#\ only in the
+topo variant}\\
\\
Plurality answer: \{plurality\}\\
\\
All agent answers:\\
\hphantom{xx}1. \{answer\_1\}\\
\hphantom{xx}2. \{answer\_2\}\\
\hphantom{xx}\ldots\\
\hphantom{xx}N. \{answer\_N\}\\
\\
Probability the plurality answer is correct (0--1):}

\vspace{6pt}\hrule\vspace{4pt}

\textbf{Topology descriptions} (used by the \texttt{+topo} variant):\\[2pt]
\texttt{iid \(\rightarrow\) "iid (independent)"\\
debate \(\rightarrow\) "debate (full-mesh cross-critique)"\\
chain \(\rightarrow\) "chain (sequential)"\\
hub\_spoke \(\rightarrow\) "hub-spoke (centralized aggregator)"\\
tree \(\rightarrow\) "tree (hierarchical aggregation by depth)"}
\end{promptbox2}

\caption{\textbf{LLM-Cal: zero-shot LLM-elicited calibration baseline.}
The model is asked to map (question, panel answers, plurality) to a
single probability that the plurality answer is correct. The
optional \texttt{+topo} variant additionally injects a one-line
description of the panel's communication topology into the user
prompt. CAGE-Cal itself does not issue any LLM prompts at inference
time.}
\label{fig:appendix:llmcal-prompt}
\end{figure*}

\section{Use of AI Assistants}
AI assistants were used only for language polishing.

\end{document}